\documentclass[letterpaper, 10 pt, conference]{ieeeconf}  

\IEEEoverridecommandlockouts                              

\usepackage[pdftex]{graphicx}
\usepackage{epstopdf} 
\usepackage{epsfig} 
\usepackage{mathptmx} 
\usepackage{times} 
\usepackage{amsmath} 
\usepackage{amssymb}  
\usepackage{subfigure}
\usepackage{mathtools}
\usepackage{physics}
\usepackage{bm}
\usepackage{adjustbox}
\usepackage[hidelinks]{hyperref}
\usepackage[capitalise]{cleveref}
\let\labelindent\relax
\usepackage[inline]{enumitem}
\usepackage{multirow}
\usepackage[usenames,dvipsnames]{xcolor}
\usepackage{algorithm}
\usepackage{algorithmic}

\Crefname{figure}{Fig.}{Figs.}

\DeclareMathOperator*{\argmin}{arg\,min}
\DeclareMathOperator{\KL}{\mathbb{K}\mathbb{L}}

\DeclareMathOperator{\EX}{\mathbb{E}}
\newcommand{\mvec}[1]{\mathbf{#1}}
\DeclarePairedDelimiterX{\infdivx}[2]{(}{)}{%
  #1\;\delimsize\|\;#2}
  \newcommand{\infdiv}{\infdivx}

\title{\LARGE \bf
Perceive, Represent, Generate: Translating Multimodal\\ Information to Robotic Motion Trajectories
}

\author{F\'abio Vital$^{*,1}$ Miguel Vasco$^{*}$ Alberto Sardinha$^{*}$ and Francisco Melo$^{*}$
\thanks{$^{*}$All authors are with INESC-ID \& Instituto Superior Técnico, University of Lisbon, Portugal}%
\thanks{$^{1}$Corresponding author: \tt\footnotesize \href{mailto:fabiovital@tecnico.ulisboa.pt}{fabiovital@tecnico.ulisboa.pt}}%
}

\begin{document}

\maketitle
\thispagestyle{empty}
\pagestyle{empty}

\begin{abstract}

We present \emph{Perceive-Represent-Generate} (PRG), a novel three-stage framework that maps perceptual information of different modalities (e.g., visual or sound), corresponding to a series of instructions, to a sequence of movements to be executed by a robot. In the first stage, we perceive and preprocess the given inputs, isolating individual commands from the complete instruction provided by a human user. In the second stage we encode the individual commands into a multimodal latent space, employing a deep generative model. Finally, in the third stage we convert the latent samples into individual trajectories and combine them into a single dynamic movement primitive, allowing its execution by a robotic manipulator. We evaluate our pipeline in the context of a novel robotic handwriting task, where the robot receives as input a word through different perceptual modalities (e.g., image, sound), and generates the corresponding motion trajectory to write it, creating coherent and high-quality handwritten words.

\end{abstract}


\section{INTRODUCTION}

Recent advancements in artificial perception~\cite{ruiz2018survey} and actuation~\cite{kroemer2021review} have fostered the widespread use of robotic systems in various tasks, such as autonomous driving~\cite{yurtsever2020survey}, industrial manufacturing~\cite{villani2018survey}, and medical~\cite{scassellati2012robots} or education~\cite{leite2013social} scenarios. Furthermore, the number of tasks that require collaboration between robots and human users is expected to increase, raising significant challenges regarding the quality of their interaction and the mismatch between their perceptual, cognitive, and actuation capabilities. To improve the efficiency of robots in such scenarios, these systems can be provided with additional sensors supplying multimodal information of its environment~\cite{Xue2020humanrobotinter}. The access to additional perceptual information is fundamental as humans often employ multiple communication channels in these scenarios, such as speech and non-verbal communication~\cite{ajoudani2018progress}.

In this work, we address the problem of \emph{how to translate multimodal commands} provided by a human user through different communication channels to a \emph{movement} executed by a robotic agent. In particular, we consider a scenario where the human user provides high-dimensional perceptual data (e.g., sound, images) related to the task, such as the words in a handwriting task. The agent's role is to decompose the raw  observations (e.g., the letter sequence forming a word) and generate the corresponding motion trajectory. Moreover, the performance of the agent must be robust to missing modality information, as the human user may not employ all possible communication channels during task execution.

To address such problem, we contribute a novel three-stage framework \emph{Perceive-Represent-Generate} (PRG) that maps multimodal perceptual information provided by a human user to a corresponding motion trajectory executed by the robot. Initially, the agent \emph{perceives} the environment, collecting and processing the raw multimodal observations into a sequence of individual task components (e.g., letters in a word). Subsequently, in the second stage, the agent \emph{represents} the individual task components, mapping them into a multimodal latent space, encoded by a deep generative model. Finally, in the third stage, the agent \emph{generates} and merges motion information decoded from the latent representations to execute the final motion.

We instantiate our PRG pipeline in a novel multimodal scenario (\emph{Robotic Dictaphone}) where the robot is provided with textual information (through a combination of sound, image, or motion observations) and generates a single motion trajectory to write the target word, mimicking human handwriting. We perform quantitative and qualitative evaluations of PRG in the \emph{Robotic Dictaphone} scenario. We start by accessing the performance of different multimodal generative models in encoding and generating information with missing modalities. In addition, we evaluate the quality of the word samples generated by the robot against human calligraphy in a large-scale user study. The results show that our approach can robustly map multimodal commands to generate accurate handwritten word samples, regardless of the set of modalities used to pass information to the agent. 

\begin{figure*}
\centerline{\subfigure[Perceive]{\includegraphics[width=0.26\textwidth]{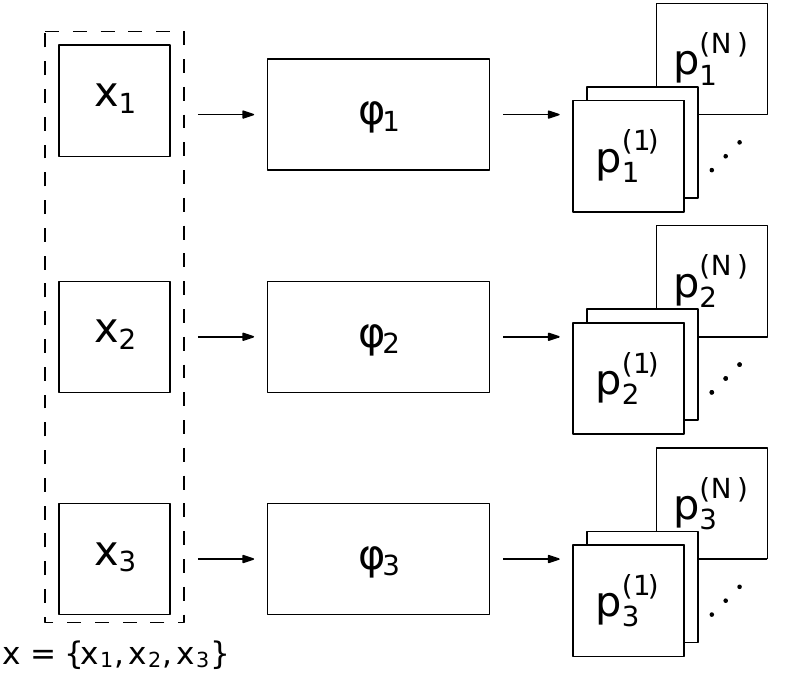}
\label{fig:prg-1}}
\hfill
\subfigure[Represent]{\includegraphics[width=0.26\textwidth]{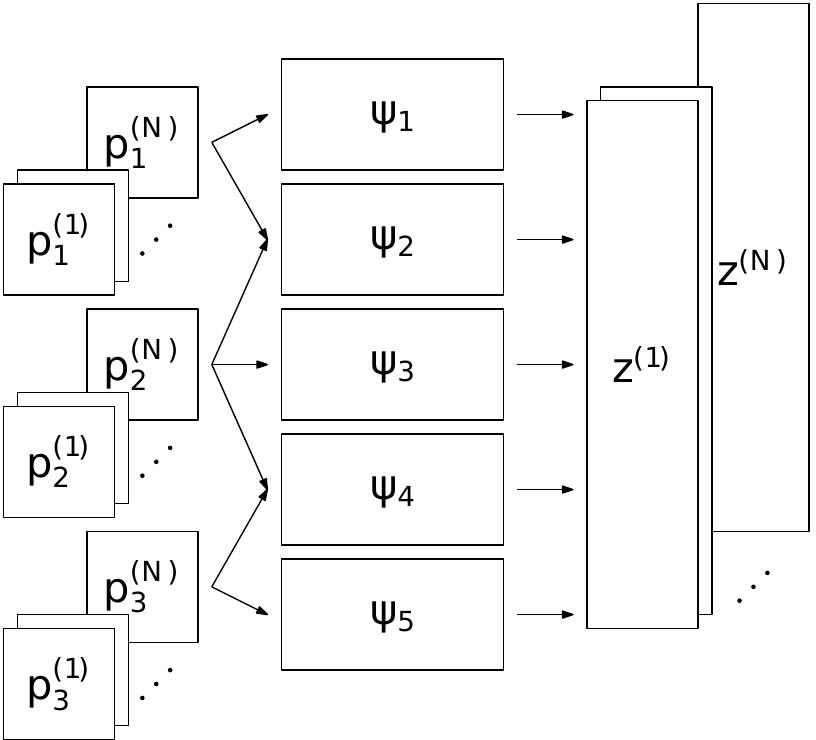}
\label{fig:prg-2}}
\hfill
\subfigure[Generate]{\includegraphics[width=0.36\textwidth]{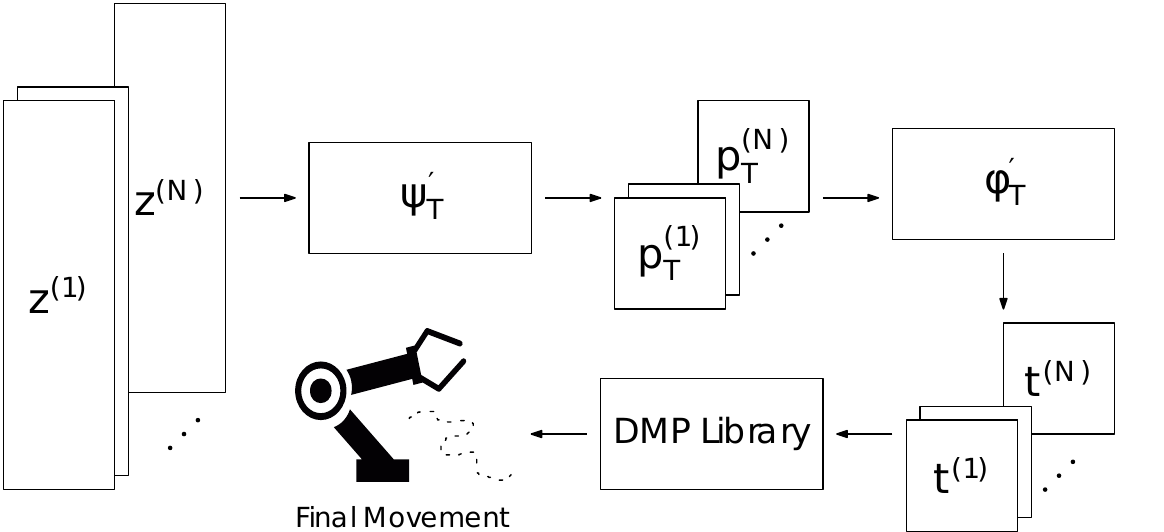}
\label{fig:prg-3}}}
\caption{The Perceive-Represent-Generate (PRG) framework for multimodal perception and actuation. Details in \Cref{section:methodology}.}
\label{fig:method-pipeline}
\end{figure*}

In summary, the main contributions of this work are:
\begin{itemize}
    \item{We propose a novel three-stage pipeline \emph{Perceive-Represent-Generate} (PRG) that translates multimodal information provided by a human user to an adequate movement executed by a robot. Crucially, such mapping is \emph{robust} to missing modality information, as the human may not always provide information through all available communication channels;}
    \item{We instantiate our PRG approach in a novel \emph{Robotic Dictaphone} scenario where textual information is converted to robotic motion trajectory, mimicking human handwriting. Our results show that, regardless of the communication channel employed by the human user (e.g., speech or image), our pipeline can accurately translate such information to generate coherent and high-quality handwritten samples.}
\end{itemize}

\section{BACKGROUND}
\label{Section:background}

In this work we employ deep generative models to encode information provided by a human user. Of relevance, the variational auto-encoder (VAE) \cite{kingma2013auto} learns to encode latent representations, \(\mvec{z}\), of high-dimensional input data, \(\mvec{x}\), without supervision. We assume that the generative process for the input data is \(\mvec{x} \sim p\left(\mvec{x}|\mvec{z}\right)\) and for the latent space is \(\mvec{z}\sim p\left(\mvec{z}\right)\), where the prior \(p\left(\mvec{z}\right)\) is usually a unit Gaussian distribution. The training of the VAE maximizes the evidence lower bound (ELBO) of the observed data \(\mvec{x}\),
\begin{equation*}
    \mathcal{L}_{\text{VAE}} = \EX_{q_{\phi}(\mvec{z} | \mvec{x})} \left[\log p_{\theta}(\mvec{x} | \mvec{z}) \right] - \KL \infdiv{q_{\phi}(\mvec{z}|\mvec{x})}{p(\mvec{z})},
    \label{Eq:background:vae_elbo_final}
\end{equation*}
where \(p_\theta\left(\mvec{x}|\mvec{z}\right)\) is a neural network parameterized by \(\theta\) (\emph{decoder}), and \(q_\phi\left(\mvec{z}|\mvec{x}\right)\) is a neural network parameterized by \(\phi\) (\emph{encoder}). A recent extension of the VAE framework to the multimodal setting is the Multimodal Unsupervised Sensing (MUSE) model~\cite{vasco2021sense}. MUSE employs an hierarchy of representations to learn modality-specific and multimodal latent representations, being robust to missing modality information and scalable to a large number of modalities.

\begin{table*}
\vspace*{5pt}
\caption{Log-likelihood metrics for different mVAE models using the augmented ``UJI Char Pen 2'' test dataset. Higher is better.}
\label{Table:log_likelihoods}
\centering
\begin{tabular}{lccccccc}
\hline
& \(\log p\left(\mvec{x}_\text{T}\right)\) & \(\log p\left(\mvec{x}_\text{S}\right)\) & \(\log p\left(\mvec{x}_\text{I}\right)\) & \(\log p\left(\mvec{x}_\text{T}|\mvec{x}_\text{S}\right)\) & \(\log p\left(\mvec{x}_\text{T}|\mvec{x}_\text{I}\right)\) & \(\log p\left(\mvec{x}_\text{S}|\mvec{x}_\text{T}\right)\) & \(\log p\left(\mvec{x}_\text{I}|\mvec{x}_\text{T}\right)\) \\ \hline
\(\text{PRG}_\text{CVAE}\left(\mvec{x}_\text{T},\mvec{x}_\text{S}\right)\) & - & - & - & -192.75 & - & - & - \\
\(\text{PRG}_\text{AVAE}\left(\mvec{x}_\text{T},\mvec{x}_\text{S}\right)\) & -197.55 & -4.17 & - & -189.28 & - & 1.94 & - \\
\(\text{PRG}_\text{AVAE}\left(\mvec{x}_\text{T},\mvec{x}_\text{I}\right)\) & -197.69 & - & -743.57 & - & -186.28 & - & -730.44 \\
\(\text{PRG}_\text{MUSE}\left(\mvec{x}_\text{T},\mvec{x}_\text{S},\mvec{x}_\text{I}\right)\) & -198.04 & -4.53 & -742.49 & -198.10 & -193.63 & 1.96 & -735.02 \\ \hline
\end{tabular}
\end{table*}

\section{METHODOLOGY}
\label{section:methodology}

We contribute with \emph{Perceive-Represent-Generate} (PRG), a novel three-stage framework that allows the encoding and generation of high-dimensional multimodal information provided by a human user. As depicted in \Cref{fig:method-pipeline}, in this work, we instantiate PRG in the context of motion trajectory generation for robotic manipulators.

\subsection{Perceive}

We assume that the robot is provided with $M$ sensors to perceive the user command, defining a perceptual space $\mathcal{X} = X_1 \times X_2 \times \ldots \times X_M$. As the user might not employ all available communication channels during task execution, the robot may not be provided with a complete command, $\mvec{x} \in \mathcal{X}$, but only with a partial view of that command.

To reduce the complexity of the high-dimensional data and remove task-irrelevant information we preprocess the input command \(\mvec{x} = \left\{ \mvec{x}_1, \ldots, \mvec{x}_M \right\}\), discarding unavailable modalities. We define $M$ perceptual maps \(\Phi = \{\phi_1, \ldots, \phi_M\}\), responsible for processing and fragmenting the available modality-specific command \(\mvec{x}_m \in X_m\) into a sequence of $N$ individual sub-commands, \({\phi_m : \mvec{x}_m \mapsto \left( \mvec{p}_{m}^{(1)},\ldots,\mvec{p}_{m}^{(N)}\right)}\). After processing each available modality-specific input, we collect, for each sub-command, the final processed data \({\mvec{p}^{(n)}=\left\{\mvec{p}_1^{(n)}, \ldots, \mvec{p}_M^{(n)}\right\}}\), where \(n \in \{1,\ldots,N\}\).

PRG is agnostic to the nature and number of the perception maps defined by the user for each specific task. Moreover, different maps can be employed to the same modality: raw sound can be encoded into a low-dimensional representation using an encoder or decomposed into label information employing a pre-trained speech-to-text model. In addition, identity perceptual maps can also be easily defined, returning the input as a sequence with one element \(\left(\mvec{p}_m^{(1)}\right)=\mvec{x}_m\).

\subsection{Represent} \label{section:methodology:represent}

In this stage, we iteratively encode the sub-commands \(\mvec{p}^{(n)}\) into a joint latent space $\mathcal{Z}$ resulting into a sequence of $N$ latent representations \(\mvec{z}^{(n)}\), where \(\mvec{z}^{(n)} \in \mathcal{Z}\). 

The encoding process employs a set of $L$ representation maps \(\Psi=\left\{\psi_1, \ldots, \psi_L\right\}\), with \(L \leq 2^{M}-1\) to consider all possible combinations of modalities. The map \({\psi_l : \left\{\mvec{p}_{l_1}^{(n)},\ldots,\mvec{p}_{l_K}^{(n)}\right\} \mapsto \mvec{z}^{(n)}}\) sequentially maps the sub-commands from the corresponding subset of available modalities into a multimodal latent representation \(\mvec{z}^{(n)}\), where \(\left\{l_1,\ldots,l_K\right\} \in \mathcal{P}\left(\left\{1,\ldots,M\right\}\right)\), \(K \leq M\), and \(\mathcal{P}\) is the powerset function. Additionally, we define $M$ generation maps \(\Psi'=\left\{\psi_1',\ldots,\psi_M'\right\}\), where each map \({\psi_m' : \mvec{z}^{(n)} \mapsto \mvec{p}_{m}^{(n)}}\) allows the generation of modality-specific data \(\mvec{p}_m^{(n)}\) from the corresponding joint latent representation \(\mvec{z}^{(n)}\). The representation and generation maps can be instantiated as the encoders and decoders, respectively, of a multimodal VAE (mVAE) model and can be learned by employing a task-specific dataset prior to task execution.

\subsection{Generate}

PRG can generate any input modality from a joint latent representation since we have a generation map (decoder) for each input modality. In this work, we instantiate PRG for the generation of motion trajectories suitable for robotic manipulators. Consequently, PRG iteratively decodes the sequence of joint latent representations \(\mvec{z}^{(n)}\) into a sequence of motion sub-commands \(\mvec{p}_\text{T}^{(n)}\) employing the target motion generation map \(\psi'_\text{T} : \mvec{z}^{(n)} \mapsto \mvec{p}_\text{T}^{(n)}\), where $\psi'_\text{T}\in\Psi'$ and ${\mvec{p}_\text{T}^{(n)}\in\mvec{p}^{(n)}}$.

Subsequently, a final processing map \({\phi'_\text{T}: \mvec{p}_\text{T}^{(n)} \mapsto \mvec{t}^{(n)}}\) is applied that, if required by the task or robotic platform, allows the transformation of the raw generated trajectories \(\mvec{p}_\text{T}^{(n)}\) into transformed motion trajectories \(\mvec{t}^{(n)}\). We find it advantageous to have this final processing map to create more complex trajectories: \(\phi'_\text{T}\) allows the transformation (e.g., sizing, translation) of each individual trajectory before merging them. Finally, all transformed motion sub-commands are concatenated and converted into a single DMP~\cite{ijspeert2013dmp} ready to be executed by the robotic agent.

\section{ROBOTIC DICTAPHONE}

We introduce the \emph{Robotic Dictaphone} scenario, where the robot's goal is to generate handwritten word samples from information provided by the human user through three different communication channels \(\mvec{x} = \{\mvec{x}_{\text{T}}, \mvec{x}_{\text{S}}, \mvec{x}_{\text{I}}\}\), where \(\mvec{x}_\text{T}\) and \(\mvec{x}_\text{I}\) are a sequence of 2D letter trajectories and letter images that compose the word, respectively, and \(\mvec{x}_{\text{S}}\) is a sound corresponding to the word. At execution time, the human user may only employ a subset of such communication channels to provide the words. Therefore, PRG must learn to encode a multimodal representation robust to potential missing modality information. We now describe each of the PRG stages for this scenario.

\begingroup
\def\arraystretch{1.2}

\begin{table*}[t]
\vspace*{5pt}
  \caption{ Trajectory samples retrieved from running $\text{PRG}_\text{MUSE} (\mvec{x}_\text{T},\mvec{x}_\text{S}, \mvec{x}_\text{I})$, in the Robotic Dictaphone scenario, when given as input the sound of the respective word, $\mvec{x}_\text{S}$ (spectrogram of the recorded sound shown), or the image of each letter of the word, $\mvec{x}_\text{I}$.}
  \label{Table:words}
  \centering
  \begin{tabular}{lcccccc}
    \hline
    & \multicolumn{2}{c}{bell} & \multicolumn{2}{c}{cat} & \multicolumn{2}{c}{jump} \\ \hline
    \multirow{5.2}{*}{Input} & Sound ($\mvec{x}_\text{S}$) & Image ($\mvec{x}_\text{I}$) & Sound ($\mvec{x}_\text{S}$) & Image ($\mvec{x}_\text{I}$) & Sound ($\mvec{x}_\text{S}$) & Image ($\mvec{x}_\text{I}$) \\
    & \adjustimage{height=1.27cm,valign=m}{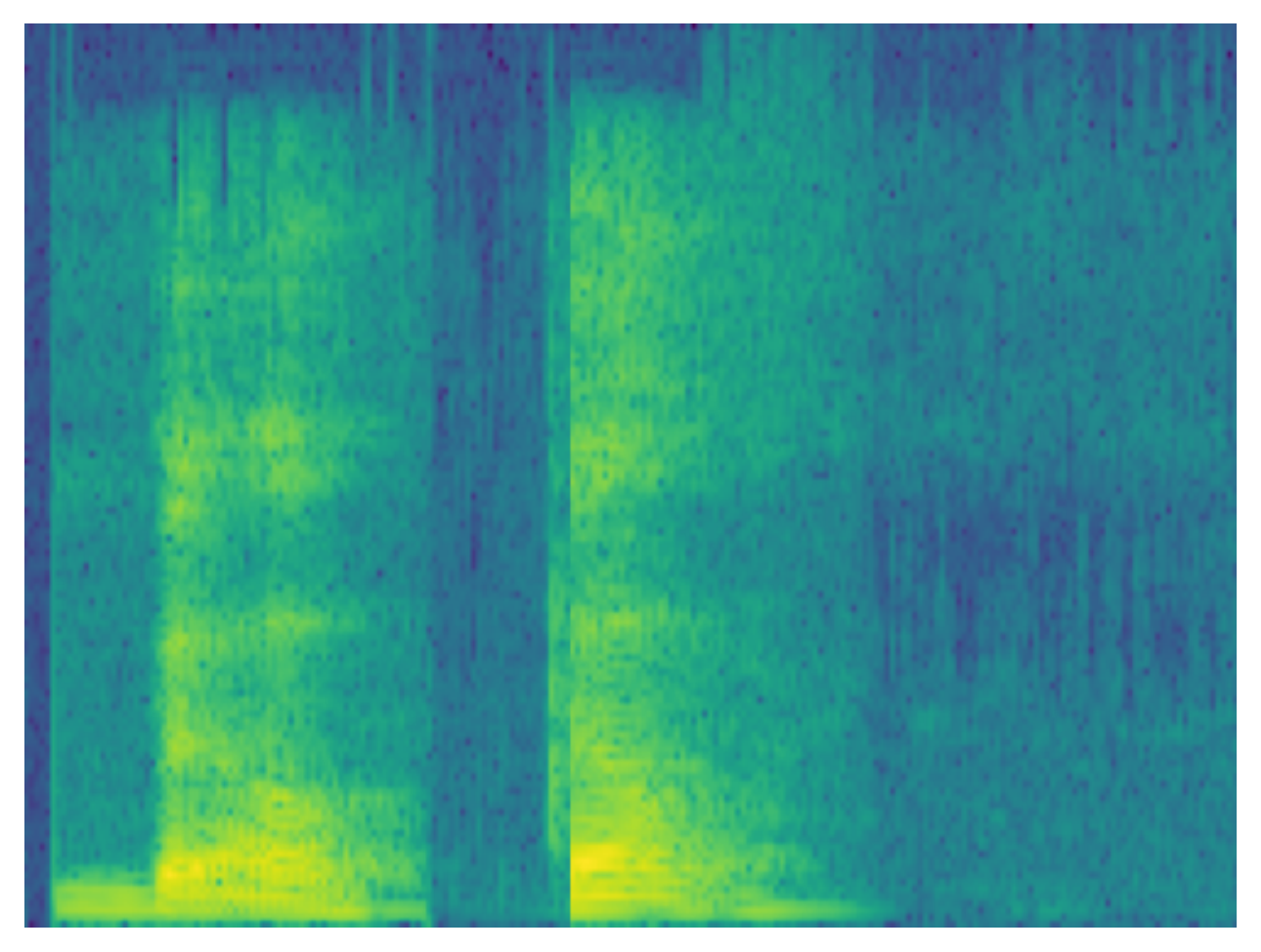} & \adjustimage{height=.83cm,valign=m}{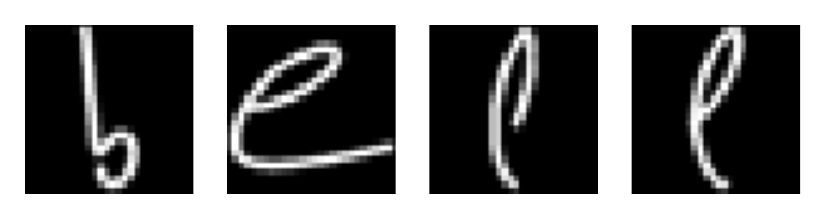} & \adjustimage{height=1.27cm,valign=m}{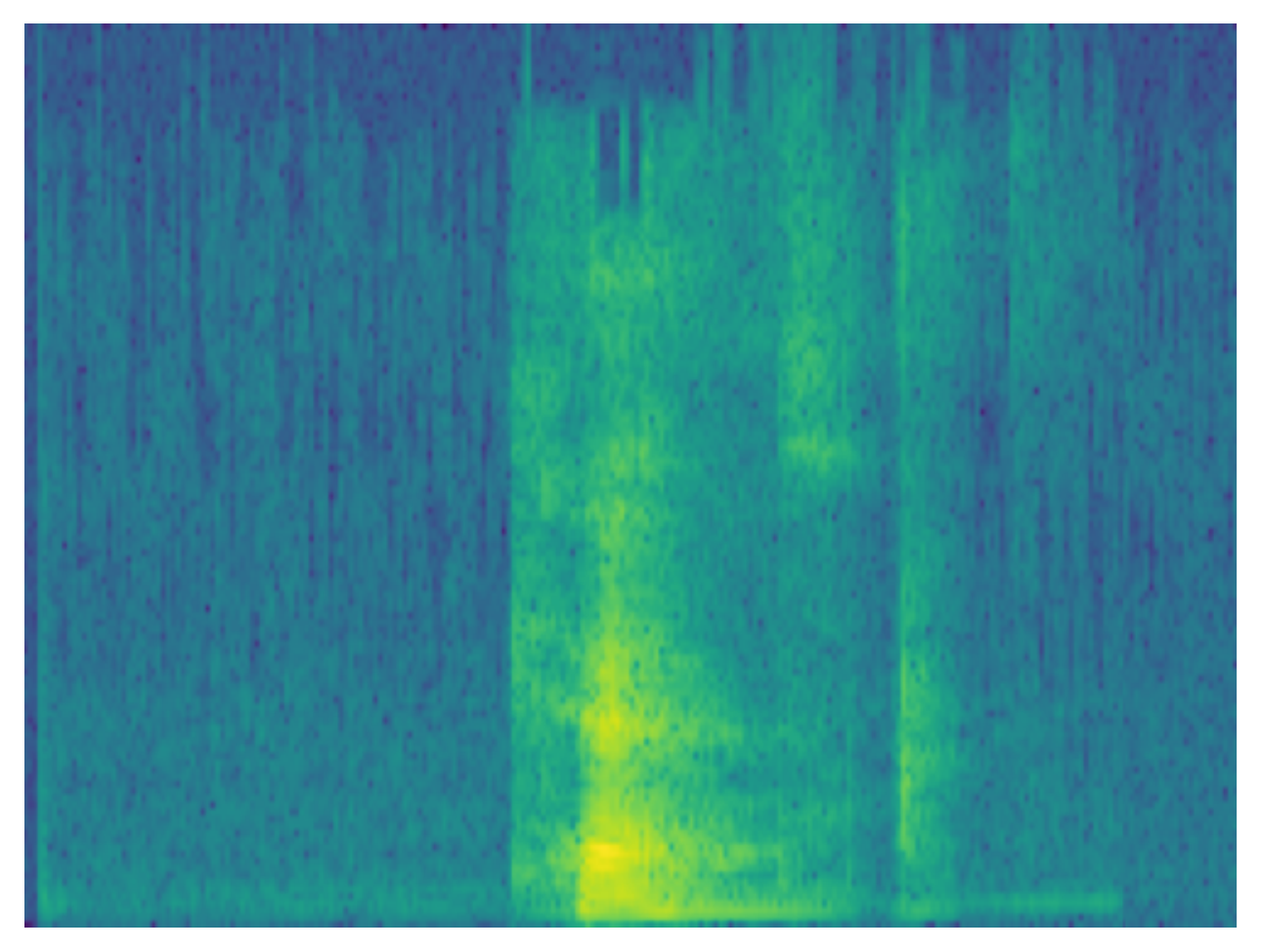} & \adjustimage{height=.83cm,valign=m}{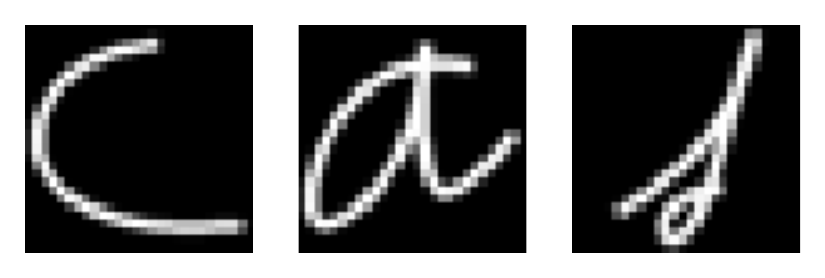} & \adjustimage{height=1.27cm,valign=m}{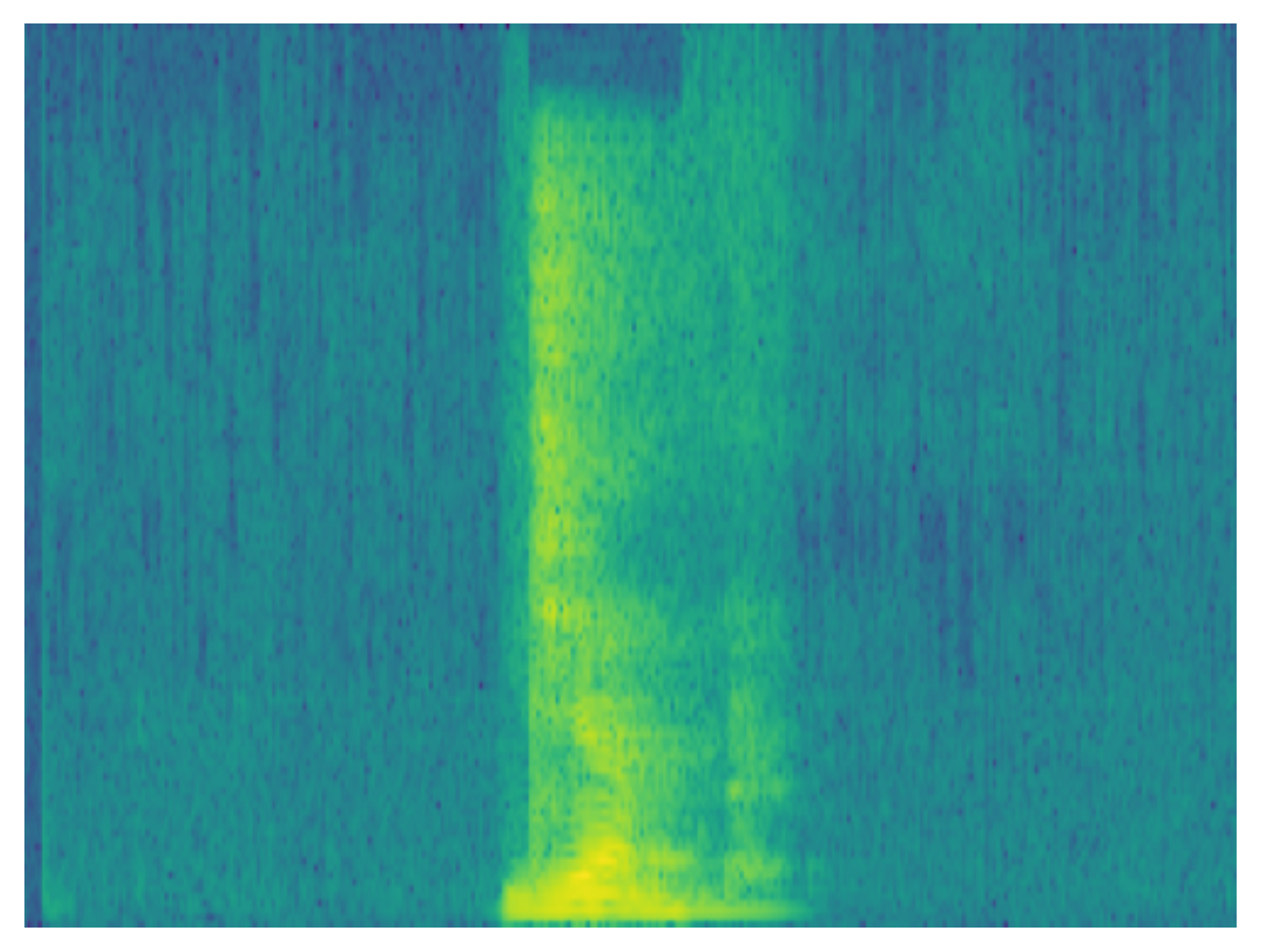} & \adjustimage{height=.83cm,valign=m}{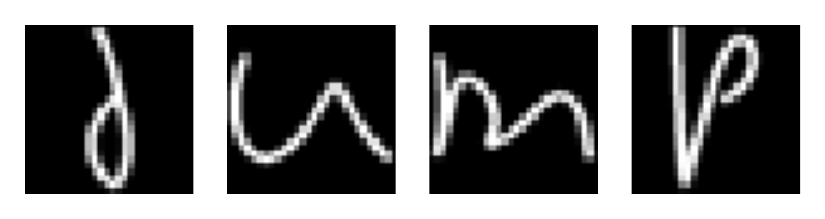} \vspace{0.ex} \\ \hline \\[\dimexpr-1.2\normalbaselineskip+0.8ex]
    Output & \adjustimage{height=.8cm,valign=m}{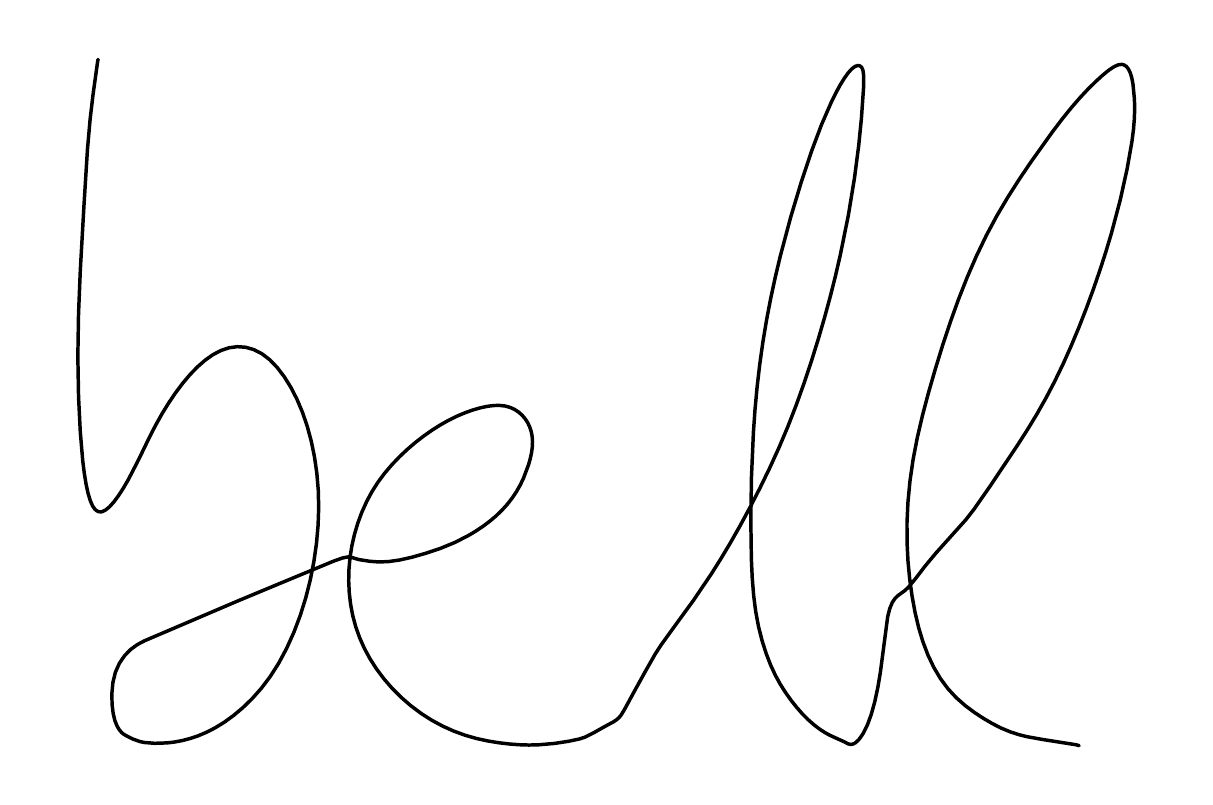} &  \adjustimage{height=.8cm,valign=m}{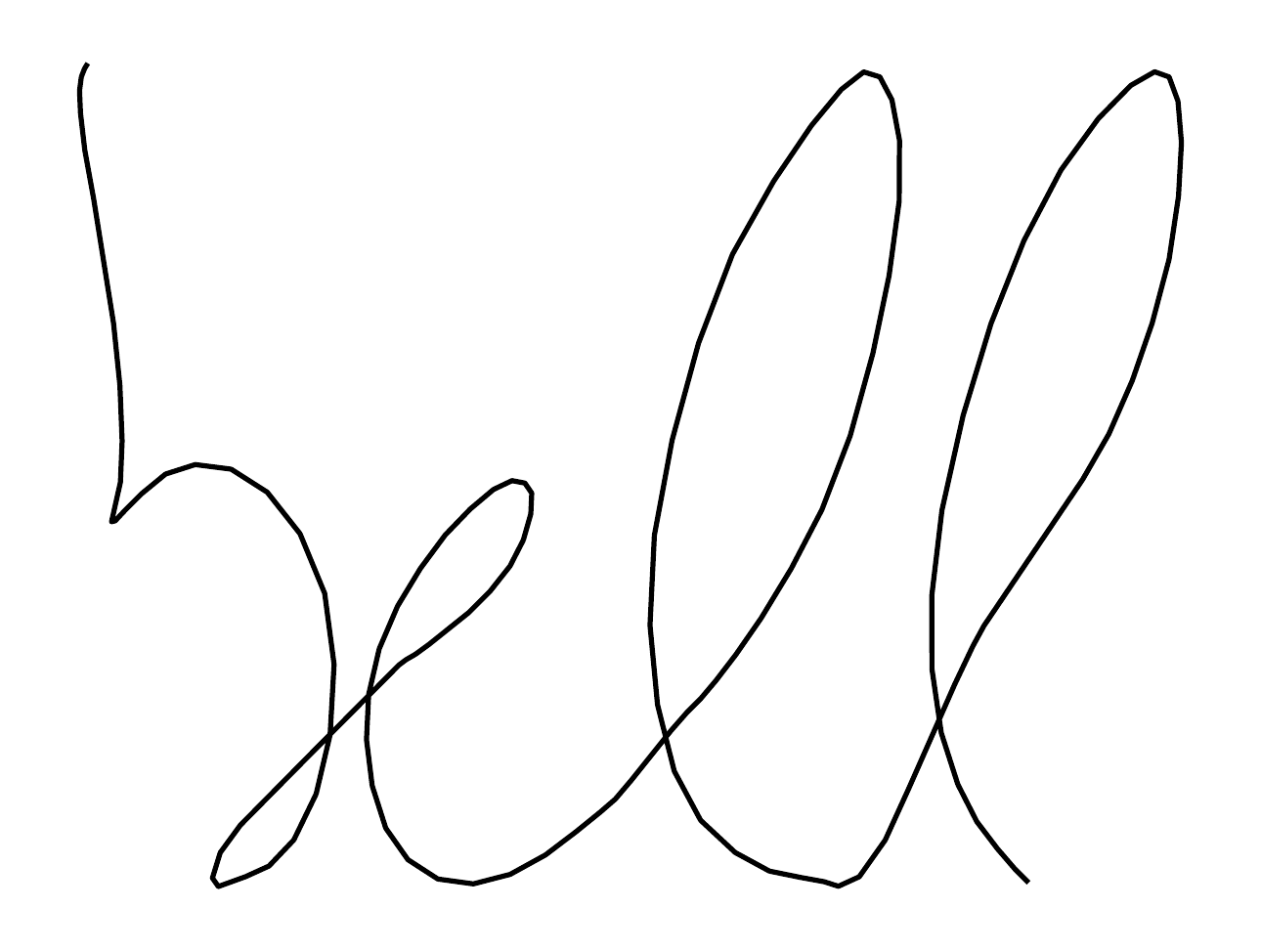} & \adjustimage{height=.8cm,valign=m}{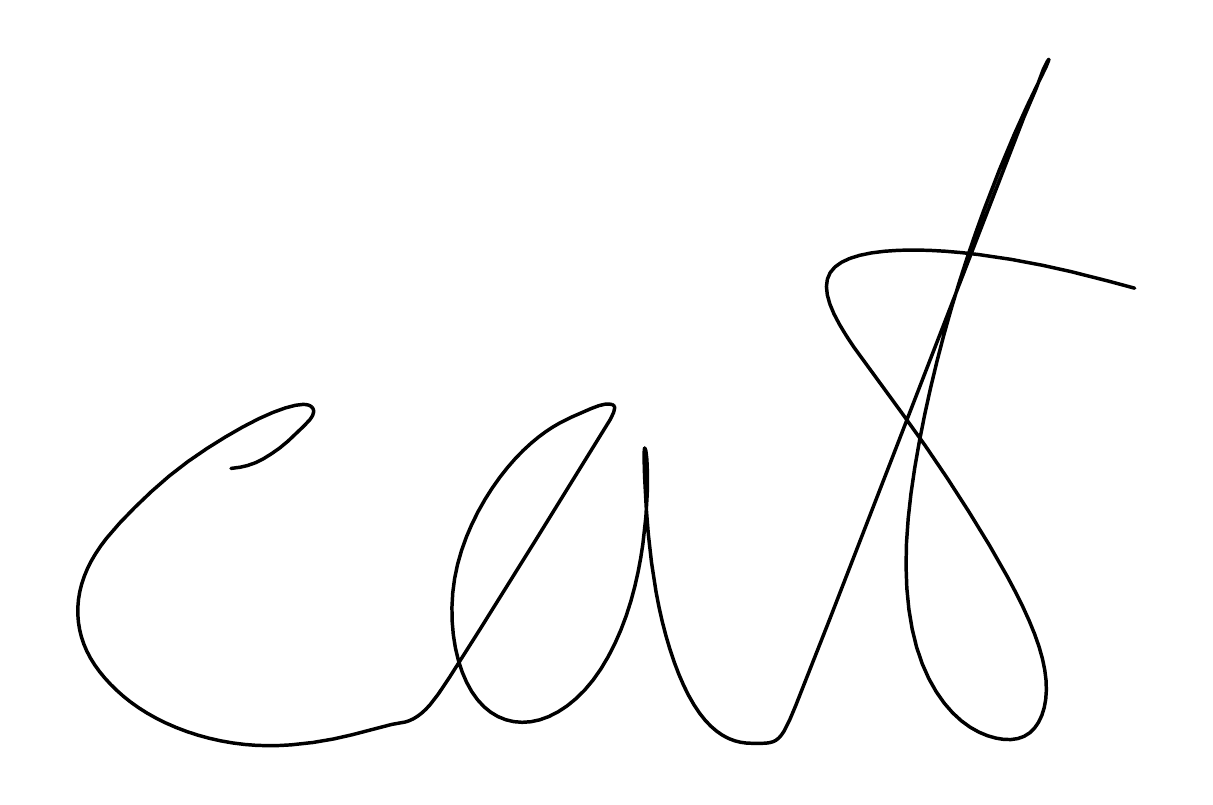} & \adjustimage{height=.8cm,valign=m}{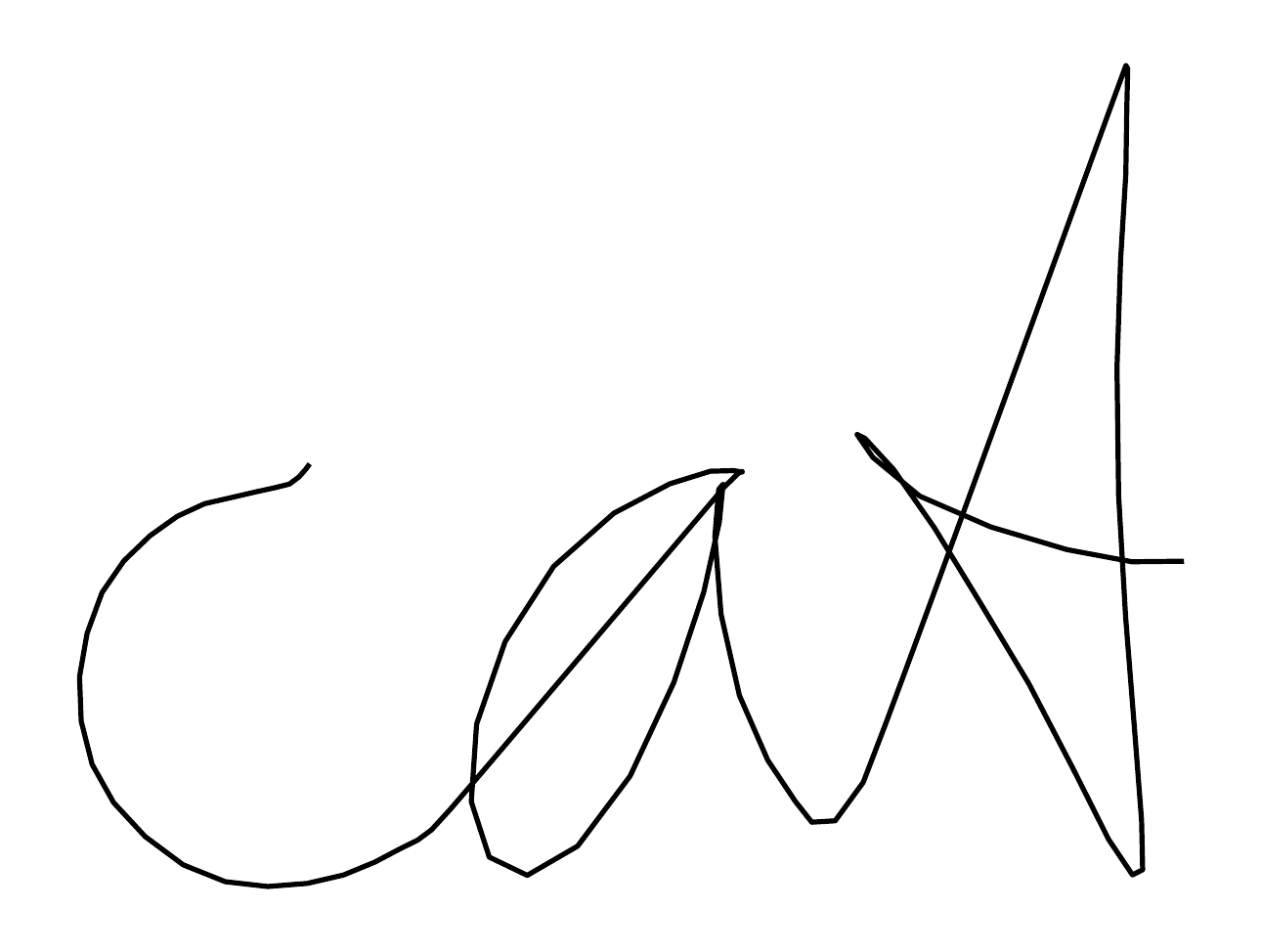} & \adjustimage{height=.8cm,valign=m}{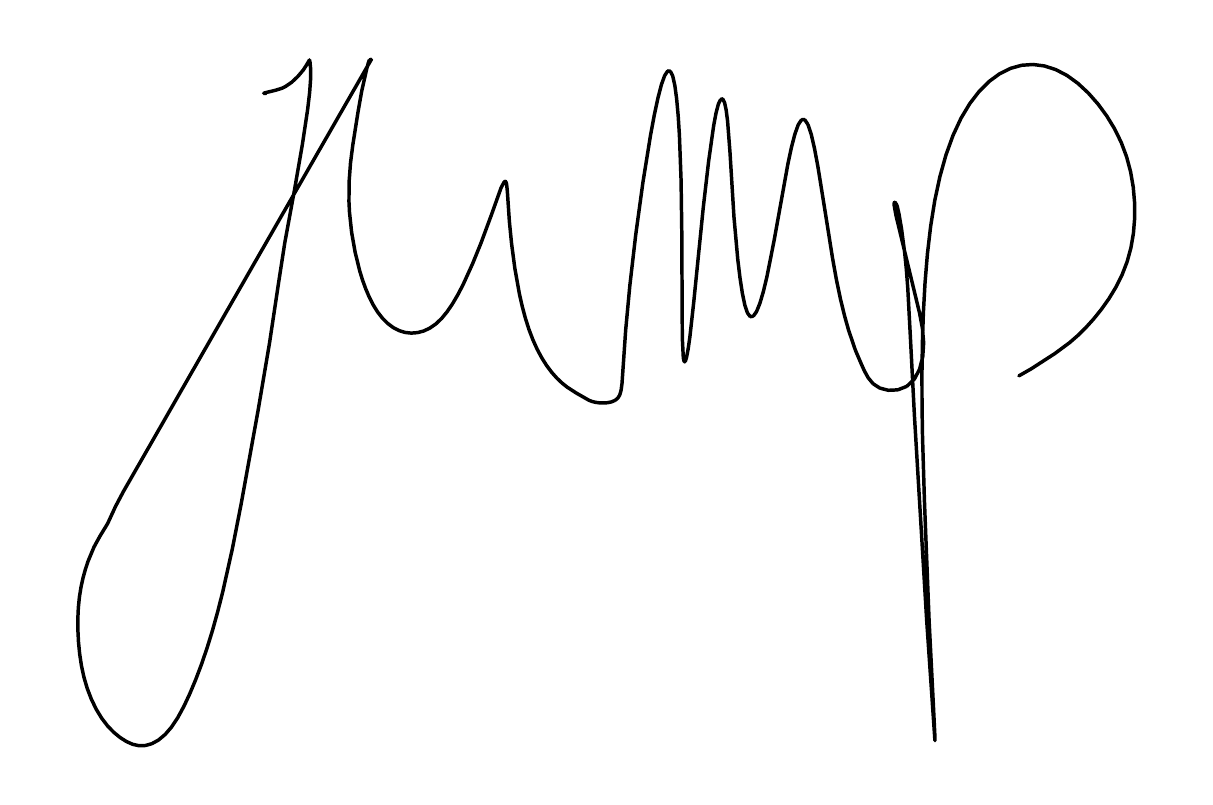} &  \adjustimage{height=.8cm,valign=m}{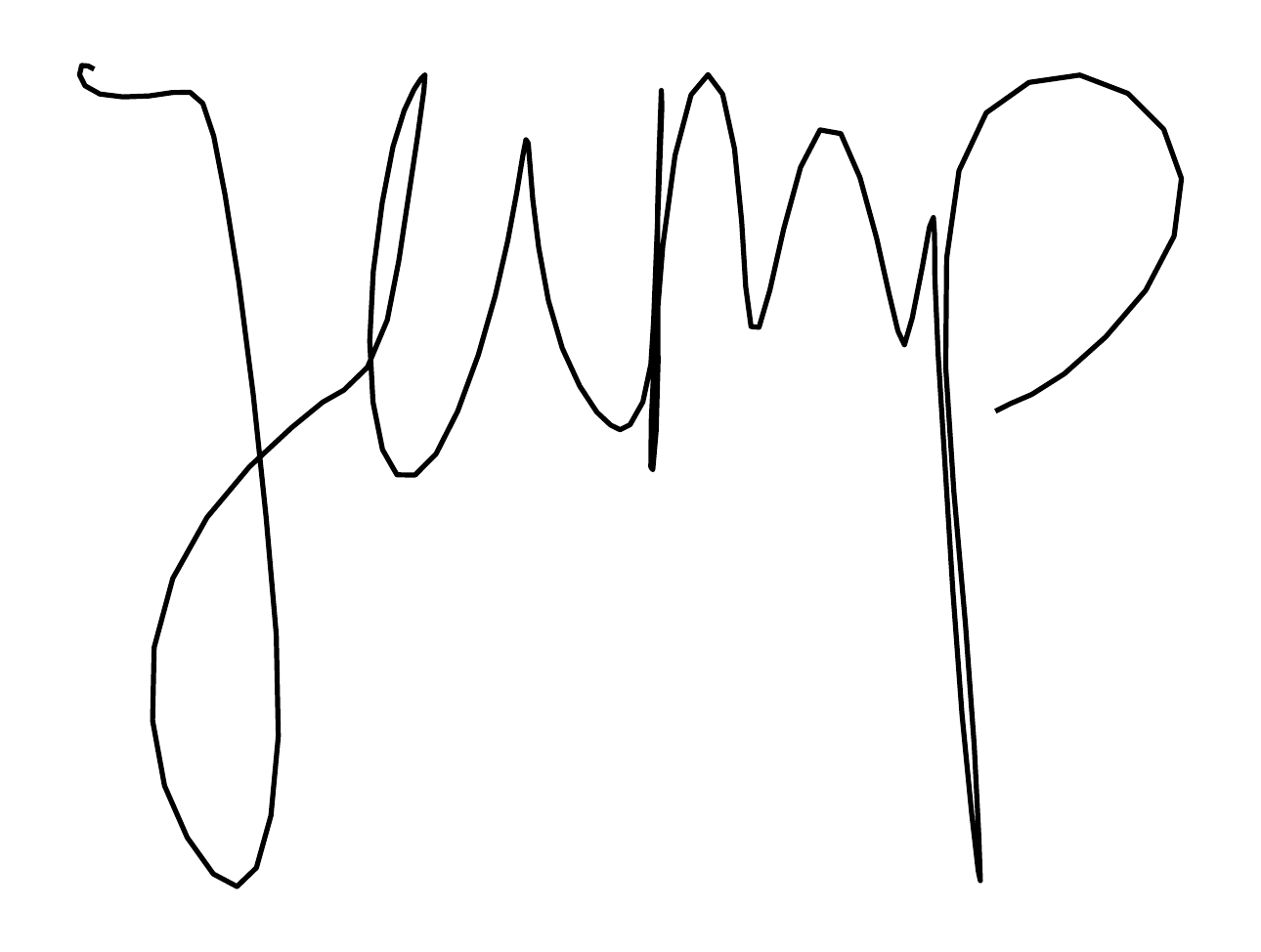} \vspace{0.8ex} \\
    \hline
\end{tabular}
\end{table*}
\endgroup

\subsection{Perceive}

The incoming raw observation data is processed and decomposed into individual sub-commands, in this case, the letters of the word. We define the perception maps specific to each modality, \(\Phi = \{\phi_\text{T}, \phi_\text{S}, \phi_\text{I}\}\). For the motion and image modalities, \(\phi_\text{T}\) and \(\phi_\text{I}\), we return the given sequence after normalizing each of its elements, trajectories and images, respectively. For the sound perception map, \(\phi_\text{S}\), we employ wav2vec 2.0, a self-supervised learning framework for speech recognition~\cite{baevski2020wav2vec}. Hence, we process the raw audio data into the label information associated with each letter, allowing for a more efficient downstream representation. 

\subsection{Represent} \label{Evaluation:Scenario:Representation}

As explained in \cref{section:methodology:represent}, PRG is agnostic to the mVAE model employed. The \emph{Represent} stage of PRG can be seen as an abstraction of mVAE's encoding phase. Depending on the mVAE model used, the PRG's representation maps will differ to accommodate the encoding channels provided by the model. 

In \Cref{Section:Eval:Quant} we evaluate the performance of PRG instantiated with different mVAE models. We train all mVAE models on data provided from the UJI Char Pen 2 dataset\footnote{To the best of our knowledge, this is the only dataset with all required modalities for English characters.}, from which we only select one-stroke formed digits and letters. We further augment the dataset by sampling from a probabilistic model derived for each character, following the procedure of~\cite{hang2016hw}.

\subsection{Generate} \label{section:robot-dict-gen}
Similarly to the \emph{Represent} stage, the PRG's \emph{Generate} stage is an abstraction of the decoding phase of the mVAE model. In particular, we consider the motion generation map \(\psi'_\text{T}\), provided by the mVAE to generate trajectory samples given the latent representation. Furthermore, we define the final processing map \(\phi'_\text{T}\) to homogenize the generated trajectories for each letter, following:
\begin{enumerate*}
    \item We scale all trajectories appropriately to their expected proportion in the final word regarding a predefined heuristic (e.g., lowercase \texttt{a} should have half-height of an uppercase \texttt{A});
    \item We translate all trajectories vertically, accordingly to the heuristic that every letter must start at the origin, except for \(\{\texttt{f},\texttt{g},\texttt{j},\texttt{p},\texttt{q},\texttt{y}\}\) which begin at a lower predefined coordinate;
    \item We define a fixed horizontal distance between two consecutive trajectories;
    \item Generate \emph{connection} trajectories\footnote{For more details regarding the final processing map refer to the extended version of the paper available at \href{https://arxiv.org/abs/2204.03051}{https://arxiv.org/abs/2204.03051}} between the end and the beginning of two consecutive letters.
\end{enumerate*}

After applying \(\phi'_\text{T}\), we concatenate the \(N\) letter trajectories and \(N-1\) connection trajectories and convert them into a single DMP for the target word, executable by the robot.

\section{Evaluation} \label{Section:Eval}
We evaluate our PRG framework in the \emph{Robotic Dictaphone} scenario. Firstly, we quantitatively evaluate the generative capability of different mVAE models integrated into PRG. Secondly, we evaluate qualitatively the handwritten samples generated by PRG considering different input modalities. Finally, we assess the quality of PRG samples against human handwriting in a large-scale user study.

\subsection{Quantitive Evaluation of mVAE models for PRG} \label{Section:Eval:Quant}
We employ and compare several mVAE models to learn the representation and generation maps $\Psi$ and $\Psi'$, respectively, required to encode multimodal data and generate the target motion trajectories. To evaluate the robustness of the \emph{Represent} and \emph{Generate} stages, we consider four different mVAE models: 
\begin{enumerate}
    \item \(\text{PRG}_\text{CVAE}(\mvec{x}_\text{T},\mvec{x}_\text{S})\): We employ the CVAE model~\cite{hristov2021learning} to learn the set of maps \(\Psi = \{\psi_{\text{T},\text{S}},\psi_\text{S}\}\) (\(\psi_{\text{T},\text{S}}\) is only used at training time) and \(\Psi'=\{\psi'_\text{T}\}\) in order to generate motion information conditioned on sound information.
    \item \(\text{PRG}_\text{AVAE}(\mvec{x}_\text{T},\mvec{x}_\text{S})\): we employ the AVAE model~\cite{yin2017associate} to learn the set of maps \(\Psi = \{\psi_\text{T},\psi_\text{S}\}\) and \(\Psi'=\{\psi'_\text{T},\psi'_\text{S}\}\) in order to encode and generate motion and sound information.
    \item \(\text{PRG}_\text{AVAE}(\mvec{x}_\text{T},\mvec{x}_\text{I})\): we employ the AVAE model to learn the set of maps \(\Psi = \{\psi_\text{T},\psi_\text{I}\}\) and \(\Psi'=\{\psi'_\text{T},\psi'_\text{I}\}\) in order to encode and generate motion and image information.
    \item \(\text{PRG}_\text{MUSE}(\mvec{x}_\text{T},\mvec{x}_\text{S},\mvec{x}_\text{I})\): we employ the MUSE model to learn a joint representation map \(\Psi=\{\psi_{\text{T},\text{S},\text{I}}\}\), robust to missing modality information, and \({\Psi'=\{\psi'_\text{T},\psi'_\text{S},\psi'_\text{I}\}}\) to generate modality-specific information.
\end{enumerate}

We evaluate the generative performance of all mVAE solutions quantitatively. In \Cref{Table:log_likelihoods}, we present standard log-likelihood metrics regarding the marginal and conditional log-likelihoods that are estimated resorting to 1000 and 5000 importance-weighted samples, respectively. The results show no significant benefit of $\text{PRG}_\text{CVAE}$, as it is outperformed by $\text{PRG}_\text{AVAE}\left(\mvec{x}_\text{T},\mvec{x}_\text{S}\right)$ regarding the conditional log-likelihood $\log p\left(\mvec{x}_\text{T}|\mvec{x}_\text{S}\right)$. As for the $\text{PRG}_\text{MUSE}$ and both $\text{PRG}_\text{AVAE}$ models, the results highlight a compromise between the generative performance and scalability of the approaches: both instances of $\text{PRG}_\text{AVAE}$ outperform $\text{PRG}_\text{MUSE}$ in terms of learning a quality trajectory representation, $\log p\left(\mvec{x}_\text{T} \right)$, and of conditionally generating trajectory information, $\log p\left(\mvec{x}_\text{T} | \mvec{x}_\text{I}  \right)$ and $\log p\left(\mvec{x}_\text{T} | \mvec{x}_\text{S}  \right)$. However, this approach is not practically extendable to more than two modalities since it needs a new encoder for each combination of modalities. On the other hand, $\text{PRG}_\text{MUSE}$ scales linearly with the number of modalities and can learn a joint representation of all modalities, suitable to generate coherent motion information. In order to be able to consider all perceptual modalities as input, we employ $\text{PRG}_\text{MUSE}$ throughout the rest of this work.

\subsection{Qualitative Evaluation of PRG Samples}
\label{Section:Eval:Generation}

We start by qualitatively evaluating the performance of PRG in generating handwritten word samples from image and sound information. In \Cref{Table:words}, we observe that $\text{PRG}_\text{MUSE}$ allows for the generation of coherent and varied word samples, regardless of the input modality employed.

\begin{figure*}
\centerline{\subfigure[]{\includegraphics[width=0.22\textwidth]{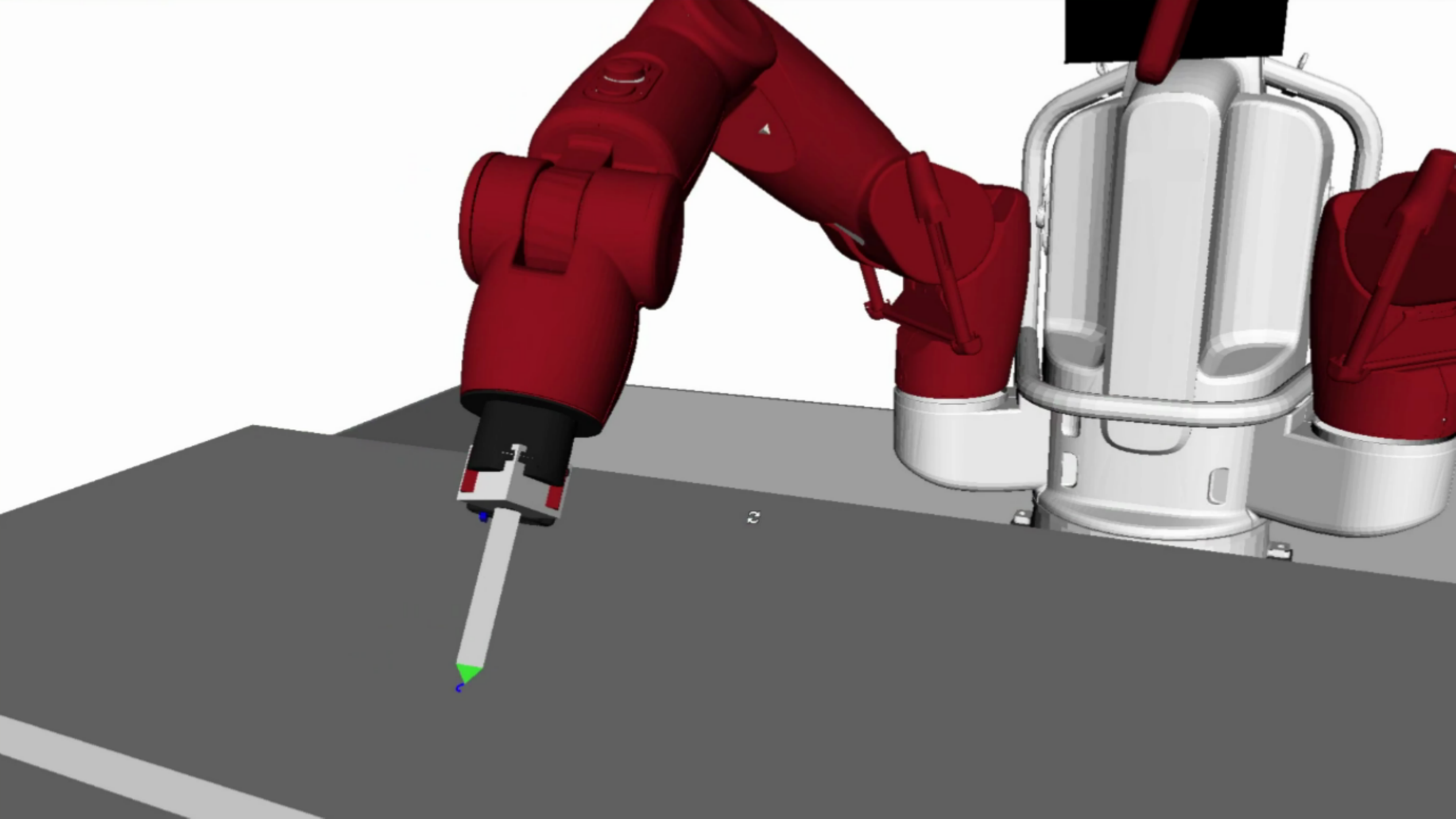}
\label{fig:baxter-2}}
\hfill
\subfigure[]{\includegraphics[width=0.22\textwidth]{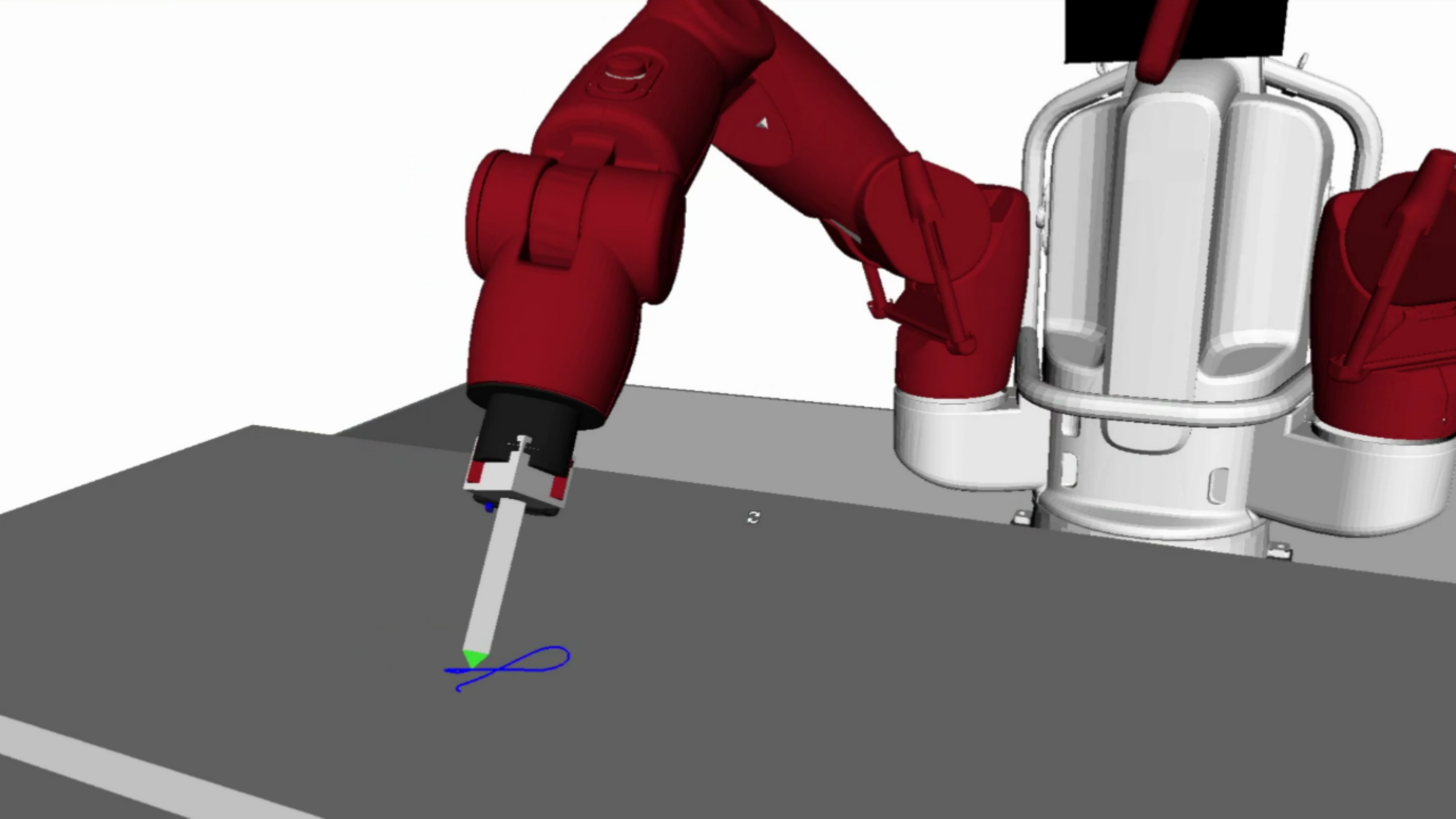}
\label{fig:baxter-5}}
\hfill
\subfigure[]{\includegraphics[width=0.22\textwidth]{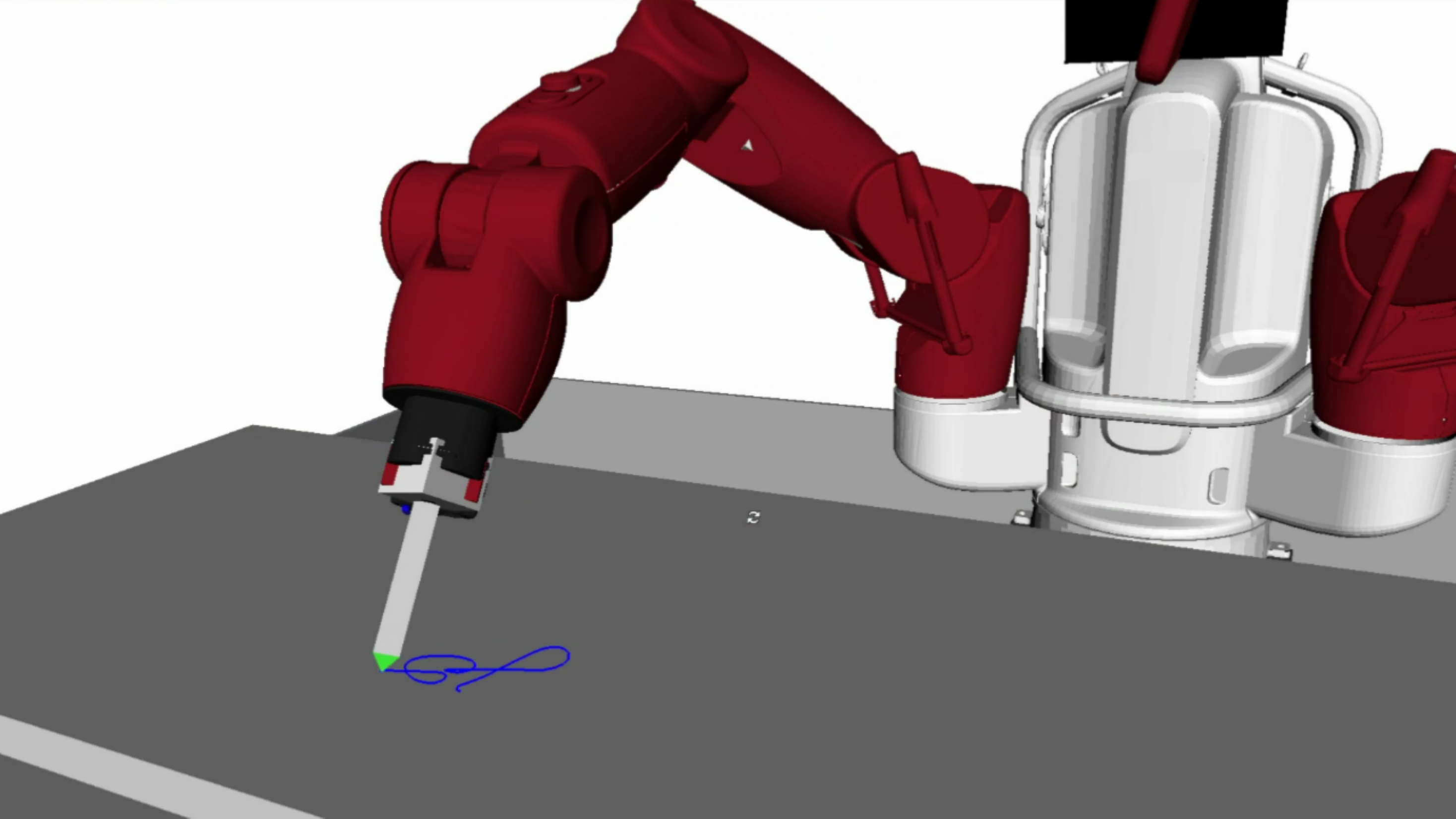}
\label{fig:baxter-8}}
\hfill
\subfigure[]{\includegraphics[width=0.22\textwidth]{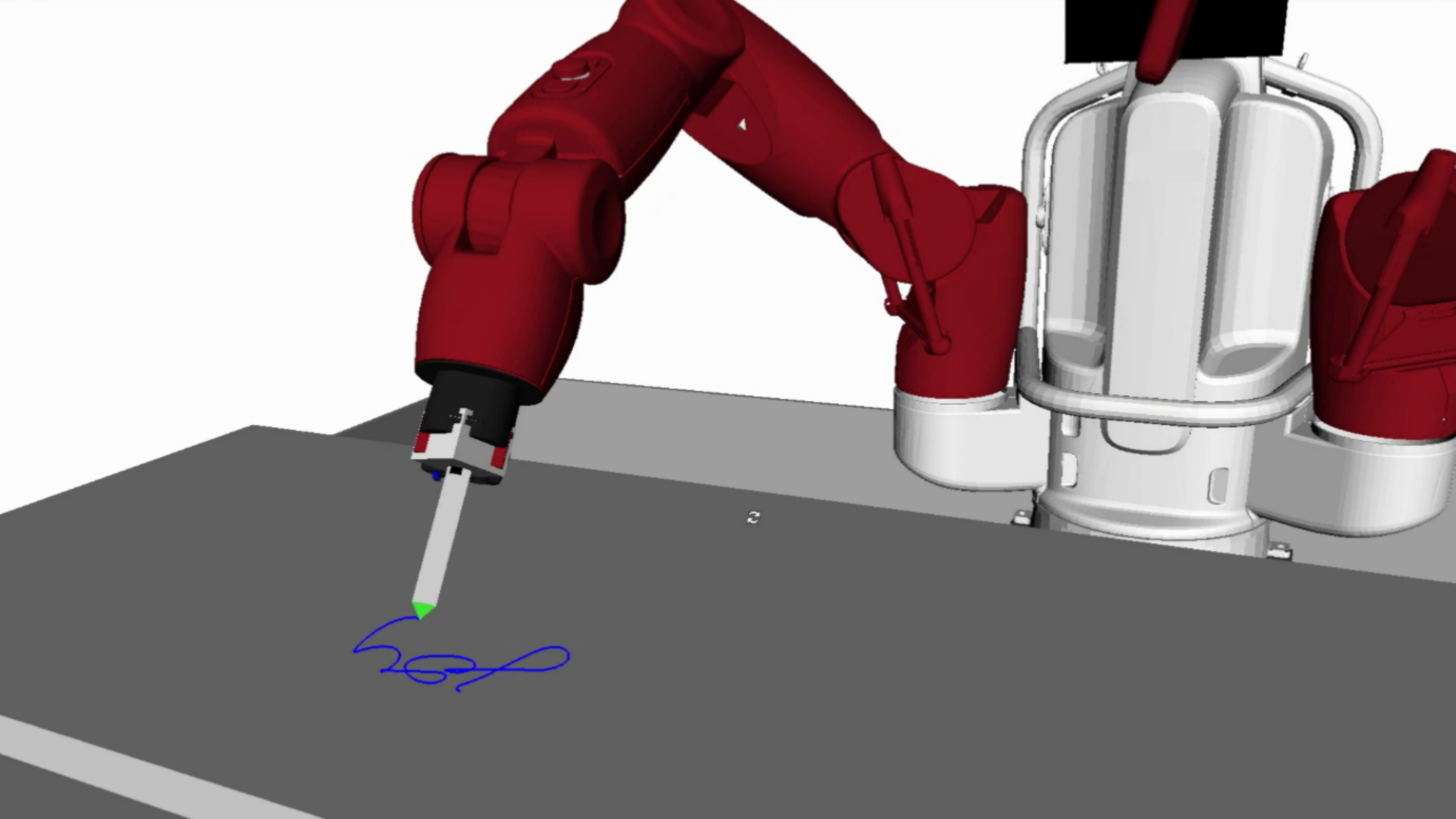}
\label{fig:baxter-10}}}
\caption{A dual 7-DOFs Baxter manipulator writing the word ``joy'' in a simulation environment on OpenRAVE. The word motion was dervived by\\ $\text{PRG}_\text{MUSE} (\mvec{x}_\text{T},\mvec{x}_\text{S}, \mvec{x}_\text{I})$ in the context of the \emph{Robotic Dictaphone} scenario.}
\label{fig:baxter}
\end{figure*}

Additionally, we highlight how the PRG framework can be incorporated into standard robotic platforms by considering a simulated environment where the robotic agent executes the final generated trajectory, provided by PRG. We use OpenRAVE~\cite{diankov2010openrave} as our simulation environment, where we place a dual 7-DOF Baxter robot in front of a table. PRG generates a single handwriting motion from an input modality of a word (\(\mvec{x}_\text{S}\) or \(\mvec{x}_\text{I}\)). The generated motion is further transformed into a joint-space trajectory using Baxter's default inverse kinematics procedure before execution. The simulation depicts how PRG allows a robot platform to convert high-dimensional inputs, such as sound, to effective motion trajectories of words, shown in \Cref{fig:baxter}.

\subsection{User Study on PRG against Human Calligraphy}

We conduct an online user study evaluating the performance of PRG in generating human-like handwritten words. The study implements a Turing-like test approach where the participants have to distinguish the origin of the word sample: human or PRG.

We start with two study hypotheses:
\begin{enumerate*}[label=(\textbf{H\arabic*})]
    \item the participant cannot distinguish motions handwritten by humans and PRG;
    \item the participant will not show high confidence when asked to distinguish handwritten words by humans and PRG.
\end{enumerate*}
With \textbf{H1}, we expect PRG to produce handwritten words similar to human handwriting: we quantify this hypothesis using $\|\hat{c}-c\|\leq\delta$, where \(\hat{c}\) and \(c\) denote the classification performance from the study and a random guess, respectively, and \(\delta\) is a threshold of equivalence. For \textbf{H2}, we expect each participant to exhibit low confidence (below the middle confidence value) in their choices, further asserting the subjective similarity of the human and PRG samples.

The study involves two phases: in a first phase, we ask 10 random participants (group 1) to write 10 words in a cursive movement (without lifting the pen). Additionally we employ PRG and generate the same words from label information. In the second phase, another group of 50 participants (group 2) answers an online and anonymous questionnaire using the Prolific platform. For each word, the participants answer two questions: in the first question, we present four randomly written words by group 1 and one by PRG, and we ask the participant to select the word written by PRG. For the second question, we ask the participant to categorize its confidence in the selection (very low, low, neutral, high, and very high).

\begin{figure}[t]
\centerline{\subfigure[]{\includegraphics[width=0.11\textwidth]{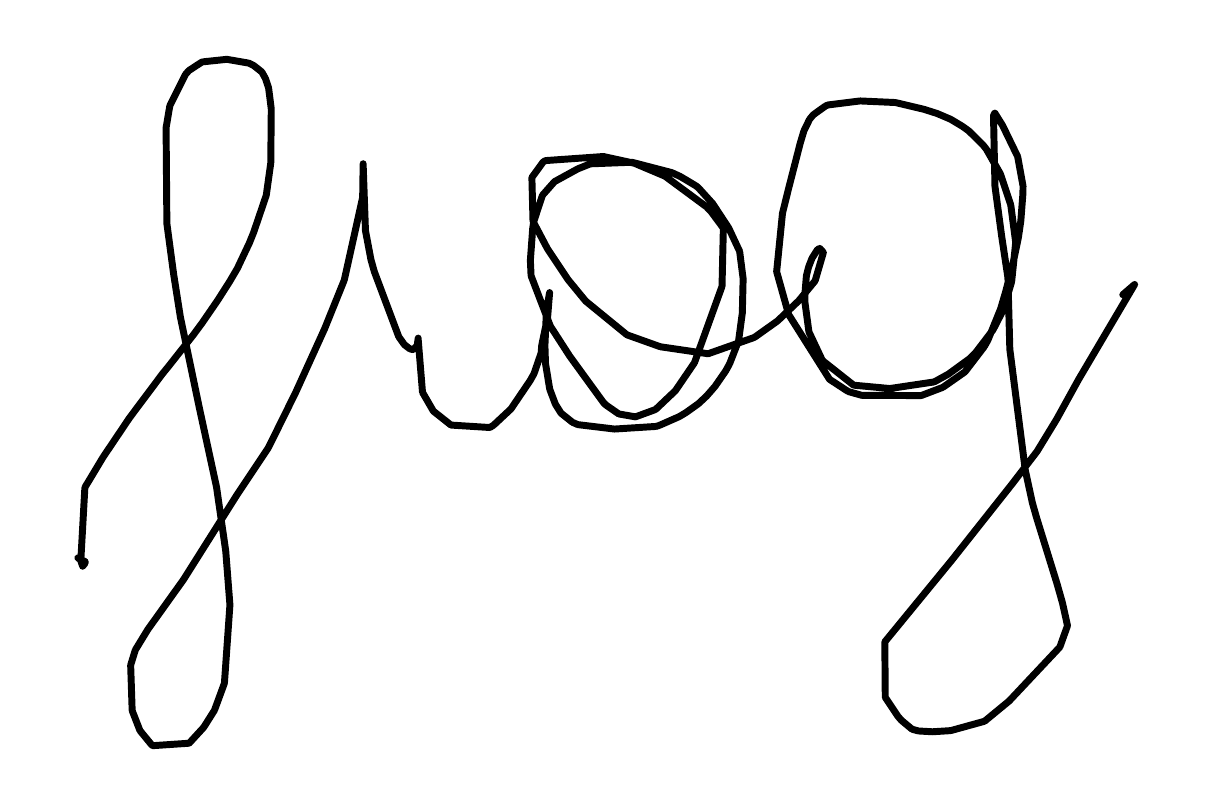}
\label{fig:user-test--user-1}}
\hfill
\subfigure[]{\includegraphics[width=0.11\textwidth]{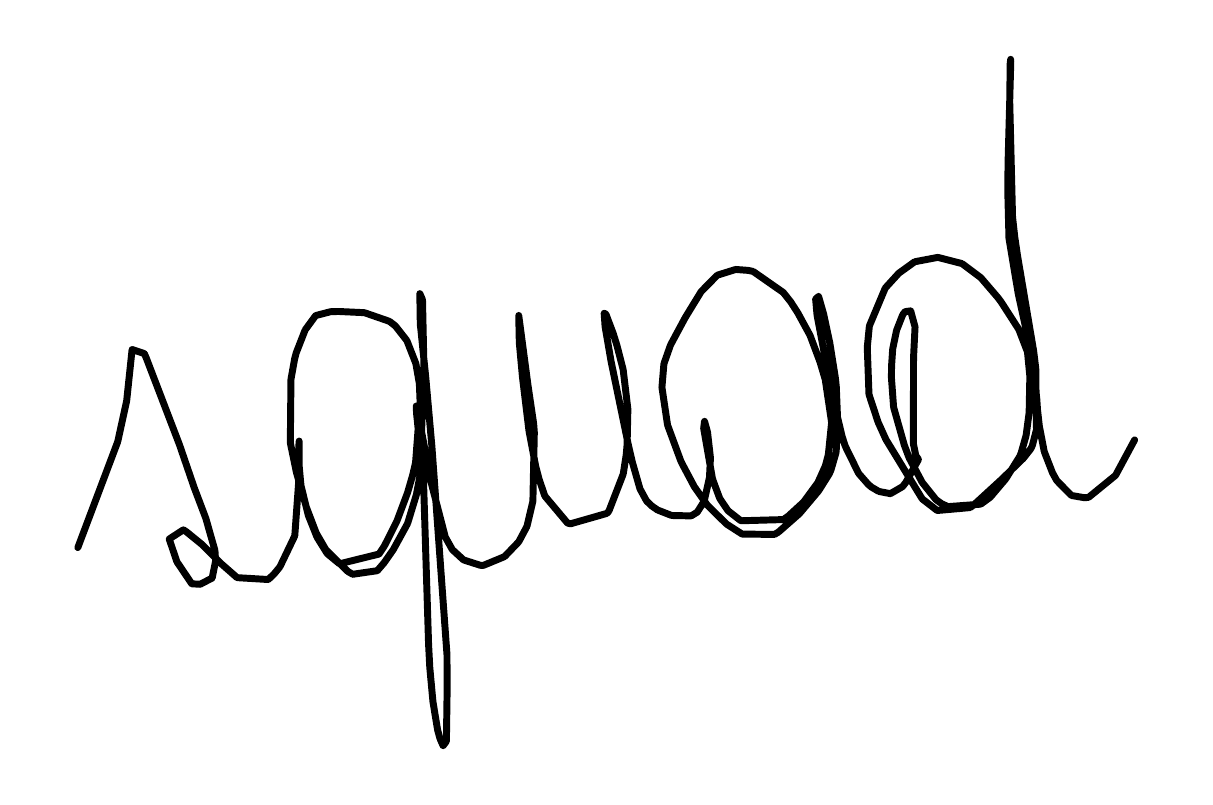}
\label{fig:user-test--user-2}}
\hfill
\subfigure[]{\includegraphics[width=0.11\textwidth]{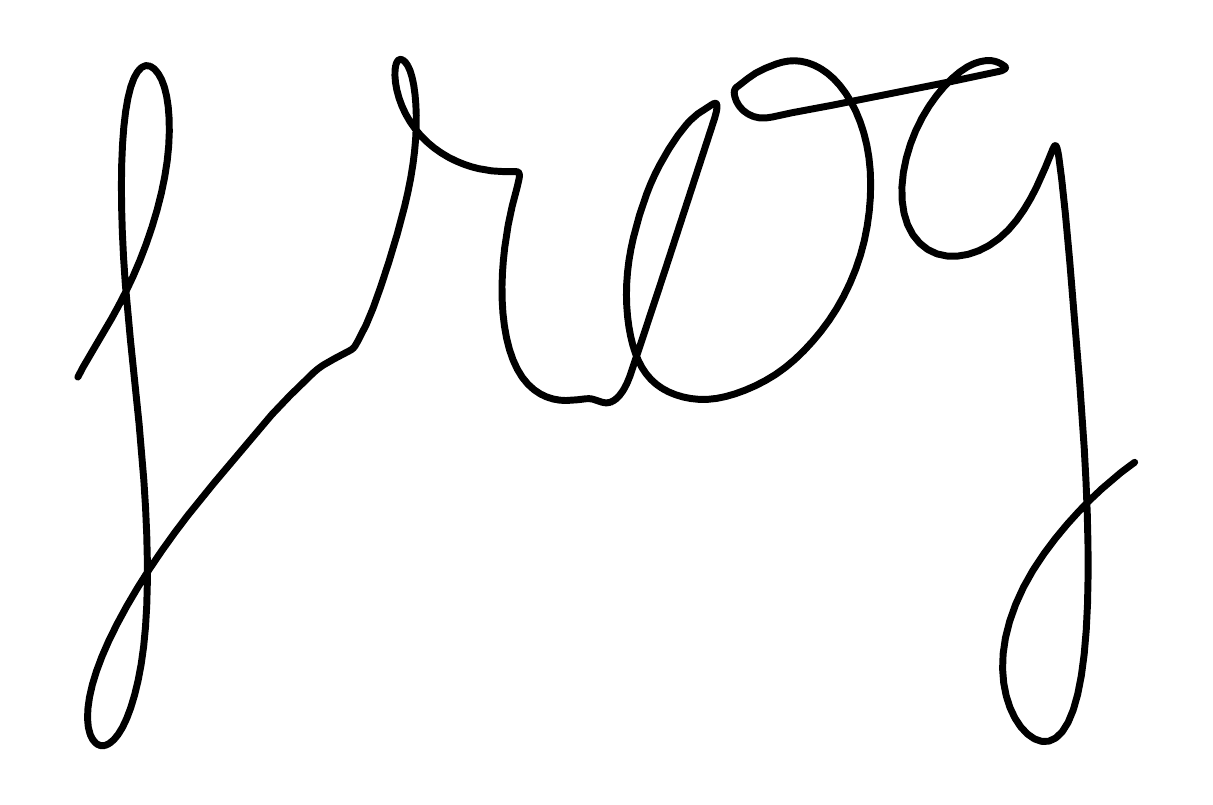}
\label{fig:user-test--prg-1}}
\hfill
\subfigure[]{\includegraphics[width=0.11\textwidth]{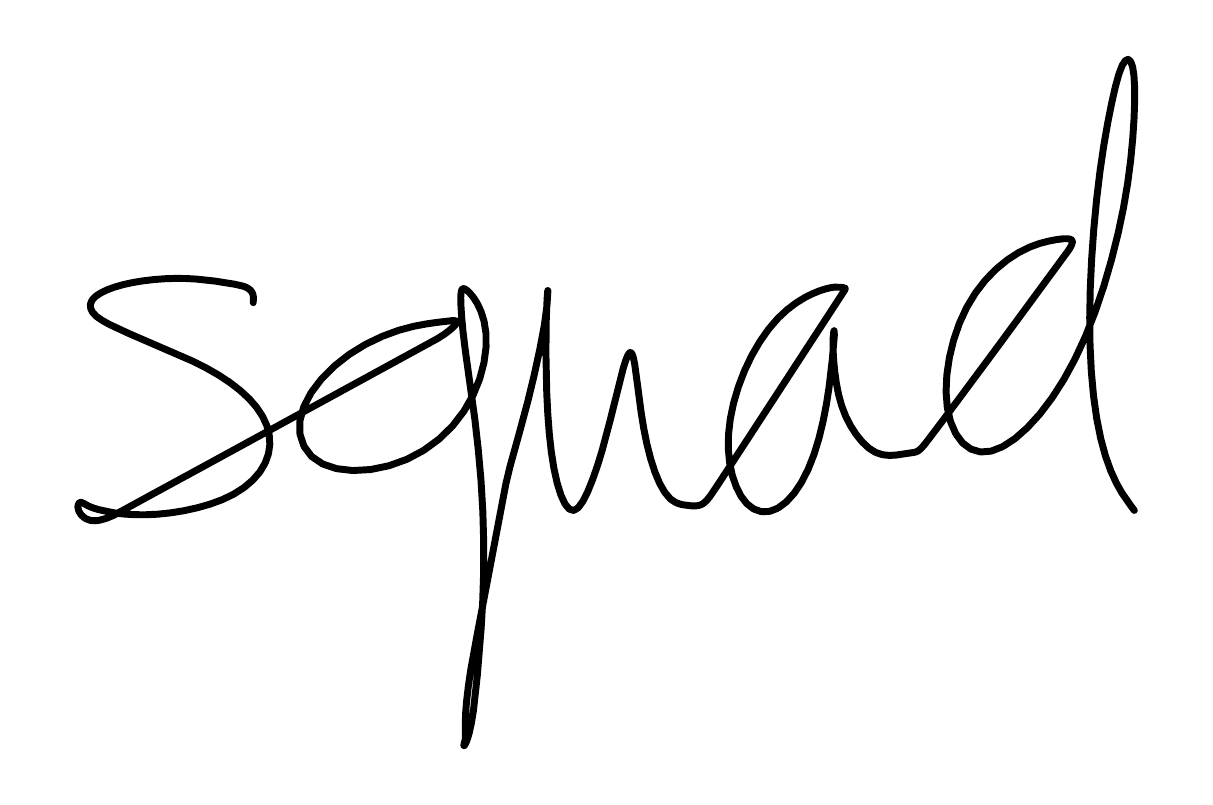}
\label{fig:user-test--prg-2}}}
\caption{Example of words handwritten by  humans \subref{fig:user-test--user-1}, \subref{fig:user-test--user-2}, and by $\text{PRG}_\text{MUSE} (\mvec{x}_\text{T},\mvec{x}_\text{S}, \mvec{x}_\text{I})$, in the context of the \emph{Robotic Dictaphone} scenario, \subref{fig:user-test--prg-1}, \subref{fig:user-test--prg-2}.}
\label{fig:user-test}
\end{figure}

In order to test \textbf{H1}, we subject the corresponding first question of each word to a binomial distribution. Our analysis showed that, on average, the participants achieved a probability of \(\hat{c} = 0.422 \pm 0.057\) for choosing the word written by PRG, which is far from a random guess, $c=0.2$. Two one-sided T-tests further supported this conclusion since it failed to reject one of the null hypotheses, in this case, \(\hat{c}<c-\delta\) where \(\delta\approx0.057\) is defined according to the confidence interval of the results obtained, with a probability threshold \(\alpha=0.05\). Such result can be understood due to the differences in appearance between the words generated by PRG and by humans: human samples appear wavier, with more fluctuations throughout the trace than the ones formed by PRG. Furthermore, as shown in \Cref{fig:user-test}, most human samples contain redundant strokes, as we forced their motion to be cursive: for example, the trace for the letter ``o'' frequently contained two full circles so the participant could adjust the motion making it easier to write the following letter. In contrast, PRG does not display this redundant behaviour since the motion of each letter is learned independently and the final word motion is converted into a single DMP.

Regarding \textbf{H2}, the average confidence level was \({3.17 \pm 1.5}\). One-sided t-test rejected that the average confidence is below neutral, 3, for \(\alpha=0.05\). The participants showed reasonable confidence about their choice, further indicating the differences between words written by humans and PRG.

To further understand the previous results, we ran a second anonymous and online user study, where we showed cursive handwritten words to a new pool of 30 participants, asking them to type the words they observed (the same from the previous study). We showed one group (half the participants) five words written by humans and the other five by PRG. For the second group (other half), we showed the same words from the opposing source, exchanging human with PRG, and vice-versa. We hypothesize that the participants are able to identify the handwritten words regardless if it was written by a human or by PRG. To test this hypothesis, we separate the results from words handwritten by humans from those by PRG. The success rates are \(0.78 \pm 0.092\) and \(0.83 \pm 0.083\), respectively. We declare that both means are equal for the null hypothesis. Using a two-sided T-test, we reject the alternative hypothesis for the threshold probability \(\alpha = 0.05\), meaning that we accept the null hypothesis. Thus, we can state that words generated by PRG are equally readable as human handwriting and the difference observed in the results of the previous user study is due to stylistic traits of the words and not fundamental, semantic, ones.

\section{RELATED WORK}

Several works focus on controlling robots with commands provided by human users through different modalities. In robotic navigation tasks, the use of directional voice commands has been explored to improve the performance of the robot~\cite{robot-navigation}. Another approach considers the uncertainty in the voice commands to facilitate learning~\cite{robot-command-sound}. However, most approaches consider a single perceptual modality, often sound, to provide commands. PRG is able to consider multiple modalities to provide commands to a robotic platform.

Other works integrate multiple modalities in order to infer the desired command. A recent work captures audio and visual samples independently, converting them into scores to combine them and determine the command from a known set~\cite{rodo2016hri}. Similarly, other approaches employ neural networks to compute the confidence scores for each possible command~\cite{liu2018hri}. These scores can only classify input modalities into predefined label commands. Meanwhile, since PRG learns a joint latent representation of all modalities, it can directly generate any modality as the output command.

Other approaches learn multimodal representations to account for commands provided through multiple input modalities. One method integrates motor and sensory time-series data (motion, image, and sound) in a fused multimodal representation, employing auto-encoder (AE) models~\cite{noda2014hri}. This framework can perform cross-modal retrieval, however, since it employs a standard AE is unable to generate novel instances. Another approach introduced a multimodal architecture for cross-modal inference on visual and bathymetric data~\cite{rao2016hri}. The framework employs a hierarchy of  denoising AEs for each modality and a mixture of restricted Boltzman machines (RBM) to learn a multimodal representation. However, training a multimodal representation through RBMs is computationally expensive and prone to divergence, requiring a meticulous model design. Contrary to both approaches, PRG can generate novel instances, through a computationally stable training mechanism (employing an mVAE).

Other approaches employ mVAEs to learn latent representations of multimodal data for robotic tasks. The CVAE model, employed as a baseline in \Cref{Section:Eval:Quant}, has been successfully integrated in the architecture of robotic agents for pouring and dabbing tasks~\cite{hristov2021learning}. The limitation of CVAE is that the generative process is fixed requiring all conditional modalities to generate the target ones. PRG employs MUSE, which, on the other hand, learns a joint latent space where the conditional and target modalities can be any set of modalities. Another approach considers the AVAE model, also employed as a baseline in \Cref{Section:Eval:Quant}, to write single letters from visual information~\cite{yin2017associate}. While AVAE can also learn a joint latent space, it needs a different encoder for each combination of modalities. In contrast, PRG employs MUSE, which increases linearly with the number of modalities.

\section{CONCLUSION}

In this work, we addressed the problem of \emph{translating} multimodal commands provided by different communication channels to a sequence of movements executed by a robotic agent. We contributed with a novel three-stage pipeline (PRG) that allows the processing, mapping, and generation of trajectory information, regardless of the communication channels employed by the human user. At the core of our pipeline, we leverage multimodal generative models to learn a low-dimensional representation of the high-dimensional data provided by the human, robust to partial observations. We instantiate our pipeline in the context of a novel multimodal robotic handwriting task called the \emph{Robotic Dictaphone} scenario. Our results show that our approach allows the generation of coherent and high-quality handwritten samples, regardless of the human user's communication channels. Furthermore, since PRG can generate any modality from the learned joint latent space, we can easily revert the robotic handwriting task (given the movement to write a word, we can use our framework to output an image or sound corresponding to the given word), as PRG is agnostic to the nature of the downstream task. Additionally, we plan on extending PRG to scenarios with large number of modalities, to test its scalability, and to consider other human-robot interaction tasks, to evaluate the role of multimodal command mapping in such scenarios.

\section*{ACKNOWLEDGEMENTS}

This work was partially supported by national funds through FCT, Funda\c{c}\~{a}o para a Ci\^{e}ncia e a Tecnologia, under projects UIDB/50021/2020 (INESC-ID multi-annual funding), PTDC/CCI-COM/7203/2020  (HOTSPOT), and PTDC/CCI-COM/5060/2021 (RELEvaNT). In addition, this work is supported by the EU project TAILOR, under GA No 952215. Miguel Vasco acknowledges the FCT PhD grant SFRH/BD/139362/2018.


\bibliographystyle{IEEEtran} 
\bibliography{IEEEabrv,refs} 

\begin{thebibliography}{10}
\providecommand{\url}[1]{#1}
\csname url@samestyle\endcsname
\providecommand{\newblock}{\relax}
\providecommand{\bibinfo}[2]{#2}
\providecommand{\BIBentrySTDinterwordspacing}{\spaceskip=0pt\relax}
\providecommand{\BIBentryALTinterwordstretchfactor}{4}
\providecommand{\BIBentryALTinterwordspacing}{\spaceskip=\fontdimen2\font plus
\BIBentryALTinterwordstretchfactor\fontdimen3\font minus
  \fontdimen4\font\relax}
\providecommand{\BIBforeignlanguage}[2]{{%
\expandafter\ifx\csname l@#1\endcsname\relax
\typeout{** WARNING: IEEEtran.bst: No hyphenation pattern has been}%
\typeout{** loaded for the language `#1'. Using the pattern for}%
\typeout{** the default language instead.}%
\else
\language=\csname l@#1\endcsname
\fi
#2}}
\providecommand{\BIBdecl}{\relax}
\BIBdecl

\bibitem{ruiz2018survey}
J.~Ruiz{-}del{-}Solar, P.~Loncomilla, and N.~Soto, ``A survey on deep learning
  methods for robot vision,'' \emph{CoRR}, vol. abs/1803.10862, 2018.

\bibitem{kroemer2021review}
O.~Kroemer, S.~Niekum, and G.~Konidaris, ``A review of robot learning for
  manipulation: Challenges, representations, and algorithms,'' \emph{Journal of
  Machine Learning Research}, vol.~22, no.~30, pp. 1--82, 2021.

\bibitem{yurtsever2020survey}
E.~Yurtsever, J.~Lambert, A.~Carballo, and K.~Takeda, ``A survey of autonomous
  driving: Common practices and emerging technologies,'' \emph{IEEE access},
  vol.~8, pp. 58\,443--58\,469, 2020.

\bibitem{villani2018survey}
V.~Villani, F.~Pini, F.~Leali, and C.~Secchi, ``Survey on human--robot
  collaboration in industrial settings: Safety, intuitive interfaces and
  applications,'' \emph{Mechatronics}, vol.~55, pp. 248--266, 2018.

\bibitem{scassellati2012robots}
B.~Scassellati, H.~Admoni, and M.~Matari{\'c}, ``Robots for use in autism
  research,'' \emph{Annual review of biomedical engineering}, vol.~14, pp.
  275--294, 2012.

\bibitem{leite2013social}
I.~Leite, C.~Martinho, and A.~Paiva, ``Social robots for long-term interaction:
  a survey,'' \emph{International Journal of Social Robotics}, vol.~5, no.~2,
  pp. 291--308, 2013.

\bibitem{Xue2020humanrobotinter}
T.~Xue, W.~Wang, J.~Ma, W.~Liu, Z.~Pan, and M.~Han, ``Progress and prospects of
  multimodal fusion methods in physical human–robot interaction: A review,''
  \emph{IEEE Sensors Journal}, vol.~20, no.~18, pp. 10\,355--10\,370, 2020.

\bibitem{ajoudani2018progress}
A.~Ajoudani, A.~M. Zanchettin, S.~Ivaldi, A.~Albu-Sch{\"a}ffer, K.~Kosuge, and
  O.~Khatib, ``Progress and prospects of the human--robot collaboration,''
  \emph{Autonomous Robots}, vol.~42, no.~5, pp. 957--975, 2018.

\bibitem{kingma2013auto}
D.~P. Kingma and M.~Welling, ``Auto-encoding variational bayes,'' \emph{arXiv
  preprint arXiv:1312.6114}, 2013.

\bibitem{vasco2021sense}
M.~Vasco, H.~Yin, F.~S. Melo, and A.~Paiva, ``How to sense the world:
  Leveraging hierarchy in multimodal perception for robust reinforcement
  learning agents,'' in \emph{Proceedings of the 21st International Conference
  on Autonomous Agents and Multiagent Systems}, ser. AAMAS '22.\hskip 1em plus
  0.5em minus 0.4em\relax Richland, SC: International Foundation for Autonomous
  Agents and Multiagent Systems, 2022, p. 1301–1309.

\bibitem{ijspeert2013dmp}
A.~J. Ijspeert, J.~Nakanishi, H.~Hoffmann, P.~Pastor, and S.~Schaal,
  ``{Dynamical Movement Primitives: Learning Attractor Models for Motor
  Behaviors},'' \emph{Neural Computation}, vol.~25, no.~2, pp. 328--373, 02
  2013.

\bibitem{baevski2020wav2vec}
A.~Baevski, Y.~Zhou, A.~Mohamed, and M.~Auli, ``wav2vec 2.0: A framework for
  self-supervised learning of speech representations,'' \emph{Advances in
  Neural Information Processing Systems}, vol.~33, pp. 12\,449--12\,460, 2020.

\bibitem{hang2016hw}
H.~Yin, P.~Alves-Oliveira, F.~S. Melo, A.~Billard, and A.~Paiva, ``Synthesizing
  robotic handwriting motion by learning from human demonstrations,'' in
  \emph{Proceedings of the Twenty-Fifth International Joint Conference on
  Artificial Intelligence}, ser. IJCAI'16, 2016, p. 3530–3537.

\bibitem{hristov2021learning}
Y.~Hristov and S.~Ramamoorthy, ``Learning from demonstration with weakly
  supervised disentanglement,'' in \emph{International Conference on Learning
  Representations}, 2021.

\bibitem{yin2017associate}
H.~Yin, F.~Melo, A.~Billard, and A.~Paiva, ``Associate latent encodings in
  learning from demonstrations,'' \emph{Proceedings of the AAAI Conference on
  Artificial Intelligence}, vol.~31, no.~1, pp. 3848--3854, Feb. 2017.

\bibitem{diankov2010openrave}
R.~Diankov, ``Automated construction of robotic manipulation programs,'' Ph.D.
  dissertation, Carnegie Mellon University, Robotics Institute, August 2010.

\bibitem{robot-navigation}
M.~V.~J. Muthugala, P.~A.~S. Srimal, and A.~B.~P. Jayasekara, ``Enhancing
  interpretation of ambiguous voice instructions based on the environment and
  the user’s intention for improved human-friendly robot navigation,''
  \emph{Applied Sciences}, vol.~7, no.~8, p. 821, 2017.

\bibitem{robot-command-sound}
M.~A. V.~J. Muthugala and A.~G. B.~P. Jayasekara, ``Enhancing human-robot
  interaction by interpreting uncertain information in navigational commands
  based on experience and environment,'' in \emph{2016 IEEE International
  Conference on Robotics and Automation (ICRA)}, 2016, pp. 2915--2921.

\bibitem{rodo2016hri}
I.~Rodomagoulakis, N.~Kardaris, V.~Pitsikalis, E.~Mavroudi, A.~Katsamanis,
  A.~Tsiami, and P.~Maragos, ``Multimodal human action recognition in assistive
  human-robot interaction,'' in \emph{2016 IEEE International Conference on
  Acoustics, Speech and Signal Processing (ICASSP)}, 2016, pp. 2702--2706.

\bibitem{liu2018hri}
H.~Liu, T.~Fang, T.~Zhou, and L.~Wang, ``Towards robust human-robot
  collaborative manufacturing: Multimodal fusion,'' \emph{IEEE Access}, vol.~6,
  pp. 74\,762--74\,771, 2018.

\bibitem{noda2014hri}
K.~Noda, H.~Arie, Y.~Suga, and T.~Ogata, ``Multimodal integration learning of
  robot behavior using deep neural networks,'' \emph{Robotics and Autonomous
  Systems}, vol.~62, no.~6, pp. 721--736, 2014.

\bibitem{rao2016hri}
D.~Rao, M.~D. Deuge, N.~Nourani–Vatani, S.~B. Williams, and O.~Pizarro,
  ``Multimodal learning and inference from visual and remotely sensed data,''
  \emph{The International Journal of Robotics Research}, vol.~36, no.~1, pp.
  24--43, 2017.

\bibitem{ujicharpen}
D.~Llorens, F.~Prat, A.~Marzal, J.~M. Vilar, M.~J. Castro, J.~C. Amengual,
  S.~Barrachina, A.~Castellanos, S.~España, J.~A. Gómez, J.~Gorbe, A.~Gordo,
  V.~Palazón, G.~Peris, R.~Ramos-Garijo, and F.~Zamora,
  ``\BIBforeignlanguage{english}{The ujipenchars database: a pen-based database
  of isolated handwritten characters},'' in
  \emph{\BIBforeignlanguage{english}{Proceedings of the Sixth International
  Conference on Language Resources and Evaluation (LREC'08)}}, N.~C.~C. Chair),
  K.~Choukri, B.~Maegaard, J.~Mariani, J.~Odijk, S.~Piperidis, and D.~Tapias,
  Eds.\hskip 1em plus 0.5em minus 0.4em\relax Marrakech, Morocco: European
  Language Resources Association (ELRA), may 2008, pp. 2647--2651.

\end{thebibliography}

\clearpage

\appendices
\onecolumn

\section{MODEL ARCHITECTURE DETAILS}
We integrate three distinct mVAE models (CVAE, AVAE, and MUSE) into PRG to test its performance in the Robotic Dictaphone scenario. We will now describe the architecture of the mVAE models employed, as well as the hyper-parameters used.

Unlike VAE, the CVAE model controls the data generated in the recognition and generative processes by introducing an extra variable, the conditional variable. A straightforward approach to model CVAE is concatenating the conditional variable to the encoder and decoder inputs.

Regarding AVAE, we have a different VAE to model each distinct combination of the input modalities. In the simplest case, when having only two modalities, we have a distinct VAE to receive each modality separately. During training, the VAE's latent spaces are associated with each other, offering the possibility to perform inference from a subset of modalities to another one. The encoders and decoders of CVAE and AVAE models, employed in \(\text{PRG}_\text{CVAE} (\mvec{x}_\text{T},\mvec{x}_\text{S})\) and \(\text{PRG}_\text{AVAE} (\mvec{x}_\text{T},\mvec{x}_\text{S})\), respectively, are multi-layer perceptron networks. \Cref{table:cvae-arch,table:avae-ms-arch} describe the architecture of both models, respectively. The AVAE model employed for \(\text{PRG}_\text{AVAE} (\mvec{x}_\text{T},\mvec{x}_\text{I})\) has a different architecture for the imaging modality, where we use convolutions to encode and decode images, see \Cref{table:avae-mi-arch}.

Finally, we present the architecture used for MUSE, which mirrors a hierarchical organization. At the lower level, the objective is to extract a latent representation specific to each modality. Accordingly, we have an encoder-decoder configuration for each modality. The architectures for the encoders and decoders are identical to the ones used in the previous models for the same modality. MUSE's high-level component comprises a single encoder-decoder architecture (both are MLPs) that combines information from modality-specific latent representations into a multimodal latent representation. The architectures for the low and high levels are present in \Cref{table:muse-low-arch,table:muse-high-arch}, respectively. We present the hyper-parameters used, for all models, in \Cref{table:hyperparams}.

\renewcommand{\arraystretch}{1.25}
\begin{table}[th]
  \caption{CVAE architecture used in \(\text{PRG}_\text{CVAE} (\mvec{x}_\text{T},\mvec{x}_\text{S})\).}
  \label{table:cvae-arch}
  \centering
  \begin{tabular}{l}
    \hline
    \textbf{Encoder} \\ \hline
    \textbf{Input: } \(\left(\mathbf{x}_\text{M},\mathbf{x}_\text{S}\right) \in \mathbb{R}^{200+62}\) \\ 
    FC 512 + LeakyReLU \\
    FC 512 + LeakyReLU \\
    FC 512 + LeakyReLU \\
    FC 16, FC 16 \\
    \textbf{Output: } \(\mathbf{\mu}\in\mathbb{R}^{16},\mathbf{\sigma}\in\mathbb{R}^{16}\) \\ \hline \\ \hline
    \textbf{Decoder} \\ \hline
    \textbf{Input: } \(\left(\mathbf{z},\mathbf{x}_\text{S}\right) \in \mathbb{R}^{16+62}\) \\ 
    FC 512 + LeakyReLU \\
    FC 512 + LeakyReLU \\
    FC 512 + LeakyReLU \\
    FC 200 \\
    \textbf{Output: } \(\mathbf{x}_\text{M}\in\mathbb{R}^{200}\) \\
    \hline
\end{tabular}
\end{table}
\renewcommand{\arraystretch}{1}

\renewcommand{\arraystretch}{1.25}
\begin{table}[t]
  \caption{AVAE architecture used in \(\text{PRG}_\text{AVAE} (\mvec{x}_\text{T},\mvec{x}_\text{S})\).}
  \label{table:avae-ms-arch}
  \centering
  \begin{tabular}{l}
    \hline
    \textbf{Encoder} (motion modality) \\ \hline
    \textbf{Input: } \(\mathbf{x}_\text{T} \in \mathbb{R}^{200}\) \\ 
    FC 512 + LeakyReLU \\
    FC 512 + LeakyReLU \\
    FC 512 + LeakyReLU \\
    FC 16, FC 16 \\
    \textbf{Output: } \(\mathbf{\mu}_\text{T}\in\mathbb{R}^{16},\mathbf{\sigma}_\text{T}\in\mathbb{R}^{16}\) \\ \hline \\ \hline
    \textbf{Decoder} (motion modality) \\ \hline
    \textbf{Input: } \(\mathbf{z}_\text{T} \in \mathbb{R}^{16}\) \\ 
    FC 512 + LeakyReLU \\
    FC 512 + LeakyReLU \\
    FC 512 + LeakyReLU \\
    FC 200 \\
    \textbf{Output: } \(\mathbf{x}_\text{M}\in\mathbb{R}^{200}\) \\
    \hline \\ \\ \hline
    \textbf{Encoder} (sound modality) \\ \hline
    \textbf{Input: } \(\mathbf{x}_\text{S} \in \mathbb{R}^{62}\) \\ 
    WordEmbed 256 \\
    FC 256 + LeakyReLU \\
    FC 256 + LeakyReLU \\
    FC 16, FC 16 \\
    \textbf{Output: } \(\mathbf{\mu}_\text{S}\in\mathbb{R}^{16},\mathbf{\sigma}_\text{S}\in\mathbb{R}^{16}\) \\ \hline \\ \hline
    \textbf{Decoder} (sound modality) \\ \hline
    \textbf{Input: } \(\mathbf{z}_\text{S} \in \mathbb{R}^{16}\) \\ 
    FC 256 + LeakyReLU \\
    FC 256 + LeakyReLU \\
    FC 256 + LeakyReLU \\
    FC 62 \\
    \textbf{Output: } \(\mathbf{x}_\text{S}\in\mathbb{R}^{62}\) \\
    \hline
\end{tabular}
\end{table}
\renewcommand{\arraystretch}{1}

\renewcommand{\arraystretch}{1.25}
\begin{table}[t]
  \caption{AVAE architecture used in \(\text{PRG}_\text{AVAE} (\mvec{x}_\text{T},\mvec{x}_\text{I})\).}
  \label{table:avae-mi-arch}
  \centering
  \begin{tabular}{l}
    \hline
    \textbf{Encoder} (motion modality) \\ \hline
    \textbf{Input: } \(\mathbf{x}_\text{T} \in \mathbb{R}^{200}\) \\ 
    FC 512 + LeakyReLU \\
    FC 512 + LeakyReLU \\
    FC 512 + LeakyReLU \\
    FC 16, FC 16 \\
    \textbf{Output: } \(\mathbf{\mu}_\text{T}\in\mathbb{R}^{16},\mathbf{\sigma}_\text{T}\in\mathbb{R}^{16}\) \\ \hline \\ \hline
    \textbf{Decoder} (motion modality) \\ \hline
    \textbf{Input: } \(\mathbf{z}_\text{T} \in \mathbb{R}^{16}\) \\ 
    FC 512 + LeakyReLU \\
    FC 512 + LeakyReLU \\
    FC 512 + LeakyReLU \\
    FC 200 \\
    \textbf{Output: } \(\mathbf{x}_\text{M}\in\mathbb{R}^{200}\) \\
    \hline \\ \\ \hline
    \textbf{Encoder} (image modality) \\ \hline
    \textbf{Input: } \(\mathbf{x}_\text{I} \in \mathbb{R}^{1\times28\times28}\) \\ 
    4 \(\times\) 4 Conv. 64, stride 2, pad 1, LeakyReLU \\
    4 \(\times\) 4 Conv. 128, stride 2, pad 1, LeakyReLU \\
    4 \(\times\) 4 Conv. 256, stride 2, pad 1, LeakyReLU \\
    FC 512 + LeakyReLU \\
    FC 16, FC 16 \\
    \textbf{Output: } \(\mathbf{\mu}_\text{I}\in\mathbb{R}^{16},\mathbf{\sigma}_\text{I}\in\mathbb{R}^{16}\) \\ \hline \\ \hline
    \textbf{Decoder} (image modality) \\ \hline
    \textbf{Input: } \(\mathbf{z}_\text{I} \in \mathbb{R}^{16}\) \\ 
    FC 512 + LeakyReLU \\
    FC 2304 + LeakyReLU \\
    4 \(\times\) 4 ConvTranspose 128, stride 2, pad 1, out\_padding 1, LeakyReLU \\
    4 \(\times\) 4 ConvTranspose 64, stride 2, pad 1, LeakyReLU \\
    4 \(\times\) 4 ConvTranspose 1, stride 2, pad 1, LeakyReLU \\
    \textbf{Output: } \(\mathbf{x}_\text{I}\in\mathbb{R}^{1\times28\times28}\) \\
    \hline
\end{tabular}
\end{table}
\renewcommand{\arraystretch}{1}

\renewcommand{\arraystretch}{1.25}
\begin{table}[t]
  \caption{MUSE's low-level architecture used in \(\text{PRG}_\text{MUSE} (\mvec{x}_\text{T},\mvec{x}_\text{S},\mvec{x}_\text{I})\).}
  \label{table:muse-low-arch}
  \centering
  \begin{tabular}{l}
    \hline
    \textbf{Encoder} (motion modality) \\ \hline
    \textbf{Input: } \(\mathbf{x}_\text{T} \in \mathbb{R}^{200}\) \\ 
    FC 512 + LeakyReLU \\
    FC 512 + LeakyReLU \\
    FC 512 + LeakyReLU \\
    FC 16, FC 16 \\
    \textbf{Output: } \(\mathbf{\mu}_\text{T}\in\mathbb{R}^{16},\mathbf{\sigma}_\text{T}\in\mathbb{R}^{16}\) \\ \hline \\ \hline
    \textbf{Decoder} (motion modality) \\ \hline
    \textbf{Input: } \(\mathbf{z}_\text{T} \in \mathbb{R}^{16}\) \\ 
    FC 512 + LeakyReLU \\
    FC 512 + LeakyReLU \\
    FC 512 + LeakyReLU \\
    FC 200 \\
    \textbf{Output: } \(\mathbf{x}_\text{M}\in\mathbb{R}^{200}\) \\
    \hline \\ \\ \hline
    \textbf{Encoder} (sound modality) \\ \hline
    \textbf{Input: } \(\mathbf{x}_\text{S} \in \mathbb{R}^{62}\) \\ 
    WordEmbed 256 \\
    FC 256 + LeakyReLU \\
    FC 256 + LeakyReLU \\
    FC 16, FC 16 \\
    \textbf{Output: } \(\mathbf{\mu}_\text{S}\in\mathbb{R}^{16},\mathbf{\sigma}_\text{S}\in\mathbb{R}^{16}\) \\ \hline \\ \hline
    \textbf{Decoder} (sound modality) \\ \hline
    \textbf{Input: } \(\mathbf{z}_\text{S} \in \mathbb{R}^{16}\) \\ 
    FC 256 + LeakyReLU \\
    FC 256 + LeakyReLU \\
    FC 256 + LeakyReLU \\
    FC 62 \\
    \textbf{Output: } \(\mathbf{x}_\text{S}\in\mathbb{R}^{62}\) \\
    \hline \\ \\ \hline
    \textbf{Encoder} (image modality) \\ \hline
    \textbf{Input: } \(\mathbf{x}_\text{I} \in \mathbb{R}^{1\times28\times28}\) \\ 
    4 \(\times\) 4 Conv. 64, stride 2, pad 1, LeakyReLU \\
    4 \(\times\) 4 Conv. 128, stride 2, pad 1, LeakyReLU \\
    4 \(\times\) 4 Conv. 256, stride 2, pad 1, LeakyReLU \\
    FC 512 + LeakyReLU \\
    FC 16, FC 16 \\
    \textbf{Output: } \(\mathbf{\mu}_\text{I}\in\mathbb{R}^{16},\mathbf{\sigma}_\text{I}\in\mathbb{R}^{16}\) \\ \hline \\ \hline
    \textbf{Decoder} (image modality) \\ \hline
    \textbf{Input: } \(\mathbf{z}_\text{I} \in \mathbb{R}^{16}\) \\ 
    FC 512 + LeakyReLU \\
    FC 2304 + LeakyReLU \\
    4 \(\times\) 4 ConvTranspose 128, stride 2, pad 1, out\_padding 1, LeakyReLU \\
    4 \(\times\) 4 ConvTranspose 64, stride 2, pad 1, LeakyReLU \\
    4 \(\times\) 4 ConvTranspose 1, stride 2, pad 1, LeakyReLU \\
    \textbf{Output: } \(\mathbf{x}_\text{I}\in\mathbb{R}^{1\times28\times28}\) \\
    \hline
\end{tabular}
\end{table}
\renewcommand{\arraystretch}{1}

\renewcommand{\arraystretch}{1.25}
\begin{table}[t]
  \caption{MUSE's high-level architecture used in \(\text{PRG}_\text{MUSE} (\mvec{x}_\text{T},\mvec{x}_\text{S},\mvec{x}_\text{I})\). \\\(\odot\) is the Hadamard product.}
  \label{table:muse-high-arch}
  \centering
  \begin{tabular}{l}
    \hline
    \textbf{Encoder} (top-level) \\ \hline
    \textbf{Input: } \(\mathbf{c}=\mathbf{z}_\text{T}\odot\mathbf{z}_\text{S}\odot\mathbf{z}_\text{I} \in \mathbb{R}^{8}\) \\ 
    FC 512 + LeakyReLU \\
    FC 512 + LeakyReLU \\
    FC 512 + LeakyReLU \\
    FC 8, FC 8 \\
    \textbf{Output: } \(\mathbf{\mu}_\pi\in\mathbb{R}^{8},\mathbf{\sigma}_\pi\in\mathbb{R}^{8}\) \\ \hline \\ \hline
    \textbf{Decoder} (top-level) \\ \hline
    \textbf{Input: } \(\mathbf{z}_\pi \in \mathbb{R}^{8}\) \\ 
    FC 512 + LeakyReLU \\
    FC 512 + LeakyReLU \\
    FC 512 + LeakyReLU \\
    FC 8 \\
    \textbf{Output: } \(\mathbf{c}\in\mathbb{R}^{8}\) \\
    \hline
\end{tabular}
\end{table}
\renewcommand{\arraystretch}{1}

\renewcommand{\arraystretch}{1.25}
\begin{table}[th]
  \caption{Hyper-parameters used for all models. Each constant is described in the original work where the model is introduced (reference shown for each constant).}
  \label{table:hyperparams}
  \centering
  \begin{tabular}{lccc}
    \hline
    \textbf{Name} & \textbf{Value (CVAE)} & \textbf{Value (AVAE)} & \textbf{Value (MUSE)}\\ \hline
    Batch size & 128 & 128 & 64 \\
    Optimizer & Adam & Adam & Adam \\
    Learning rate & \(1\times10^{-4}\) & \(1\times10^{-4}\) & \(1\times10^{-4}\) \\
    \(\alpha\), see \cite{yin2017associate} & - & 1 & - \\
    \(\alpha_\text{T}\), see \cite{vasco2021sense} & - & - & 1 \\
    \(\alpha_\text{S}\), see \cite{vasco2021sense} & - & - & 1 \\
    \(\alpha_\text{I}\), see \cite{vasco2021sense} & - & - & 1 \\
    \(\lambda_\text{T}\), see \cite{vasco2021sense} & - & - & 1 \\
    \(\lambda_\text{S}\), see \cite{vasco2021sense} & - & - & 50 \\
    \(\lambda_\text{I}\), see \cite{vasco2021sense} & - & - & 1 \\
    \(\delta\), see \cite{vasco2021sense} & - & - & 1 \\
    \(\gamma_\text{T}\), see \cite{vasco2021sense} & - & - & 10 \\
    \(\gamma_\text{S}\), see \cite{vasco2021sense} & - & - & 10 \\
    \(\gamma_\text{I}\), see \cite{vasco2021sense} & - & - & 10 \\
    \(\beta\), see \cite{vasco2021sense} & - & - & 1 \\
    \hline
\end{tabular}
\end{table}
\renewcommand{\arraystretch}{1}

\section{UJI PEN CHARACTER 2 DATASET}
UJI Pen Character (version 2) is a dataset containing handwritten trajectories of all letters in the English alphabet (lower and uppercase), all numerical digits, and some other characters~\cite{ujicharpen}. Since the original dataset is very limited in the number of samples, we augment the dataset by learning a probabilistic model derived for each character and then resampling with various constraints. For more details on the augmentation procedure, see~\cite{hang2016hw}.

After augmenting the dataset, we get more than 70000 samples, having around 1000 samples per character. Each sample has three modalities:
\begin{enumerate}
    \item Motion trajectory is a 2D Cartesian trajectory with a length of 100 equidistant points.
    \item Image derived from the sample's motion trajectory, where we construct a grayscale image with 28x28 pixels.
    \item Sample's class is a number between 0 and 61 to consider all digits and English letters (we discriminate between uppercase and lowercase letters).
\end{enumerate}

\section{Final Processing Map}
In \Cref{section:robot-dict-gen}, we summarize the procedure of PRG's final processing map, \(\phi'_\text{T}\), implemented for the Robotic Dictaphone scenario. We now detail the algorithm used in the fourth and final step of the map. Our algorithm is essential to create regular and eligible handwritten motions. Using only straight lines to connect consecutive letter trajectories turns the final motion unnatural and hard to read, where we have abrupt changes when transitioning between succeeding letter trajectories.

\begin{figure*}[th]
    \centering
    \subfigure[]{\includegraphics[width=0.49\linewidth]{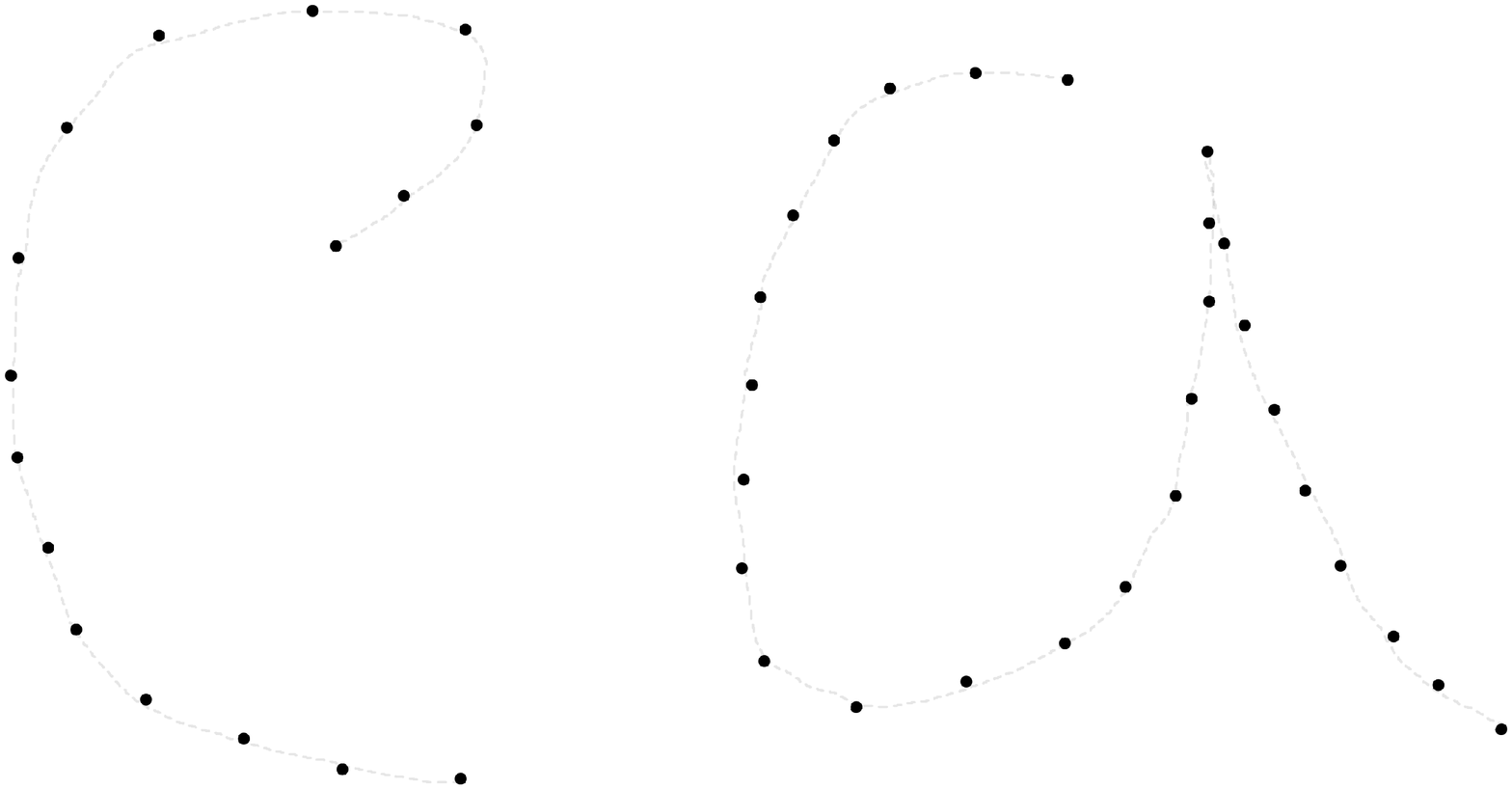} \label{fig:alg-connect-1}} \hfill
    \subfigure[]{\includegraphics[width=0.49\linewidth]{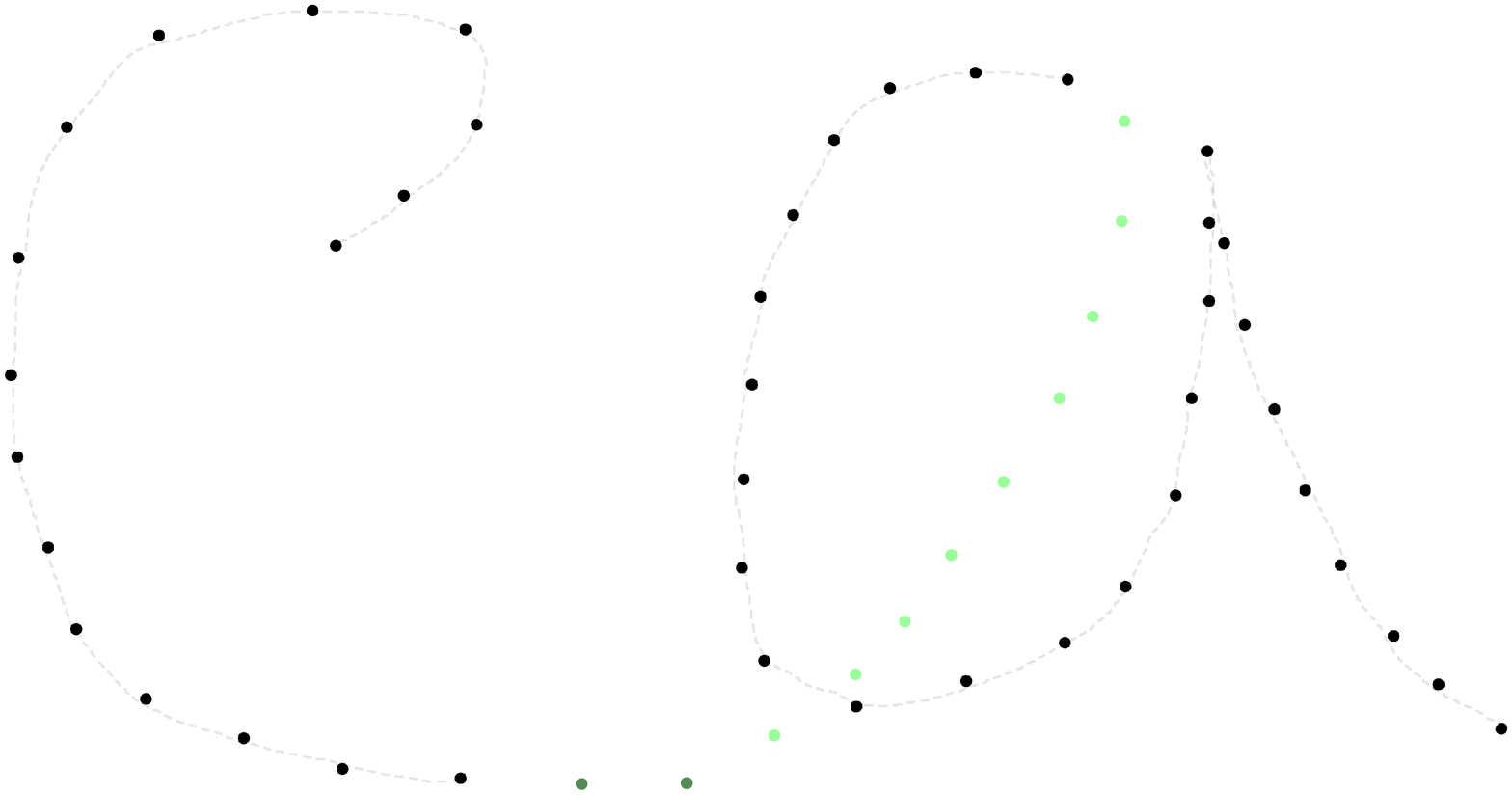} \label{fig:alg-connect-2}}
    \subfigure[]{\includegraphics[width=0.49\linewidth]{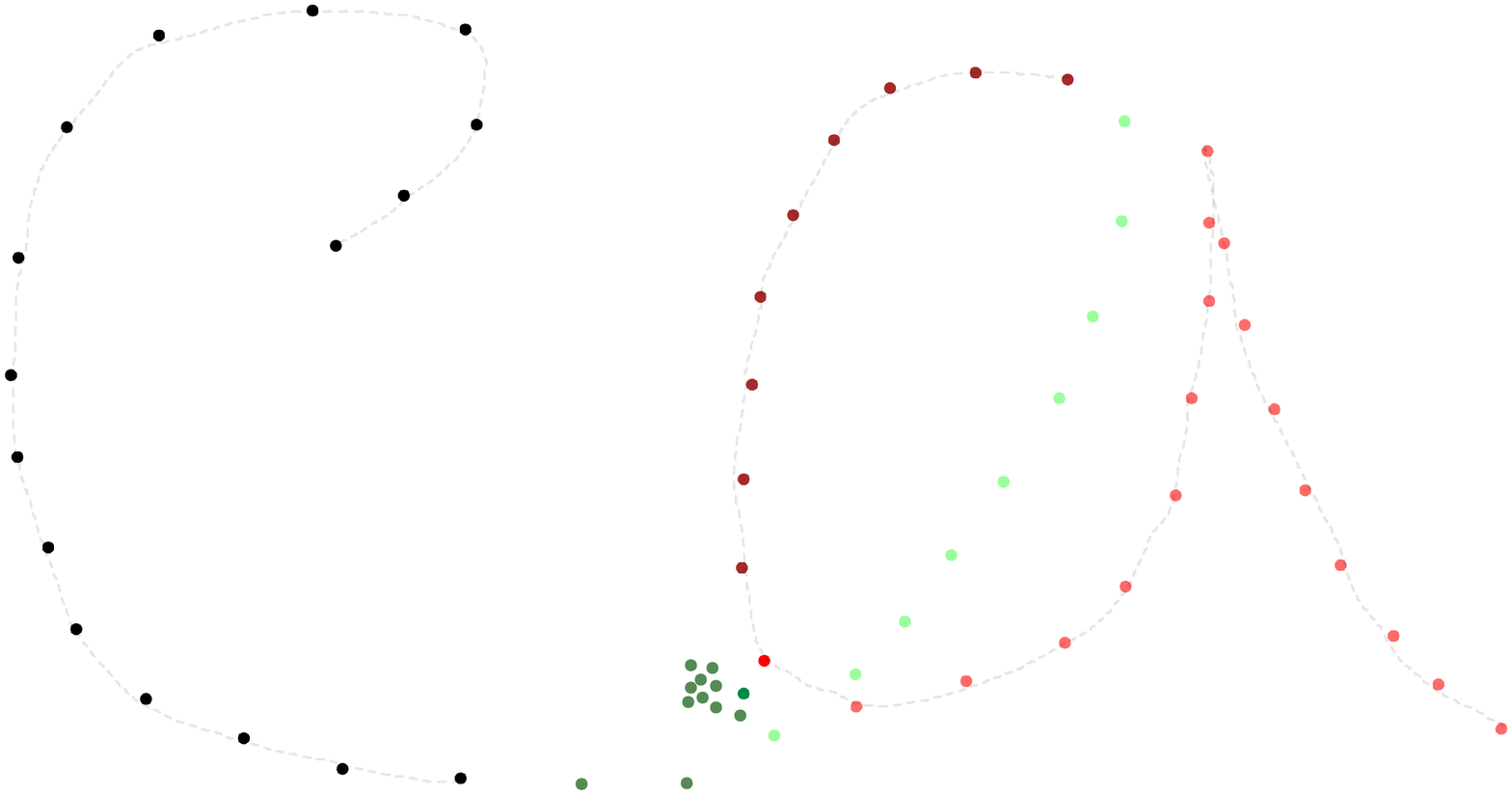} \label{fig:alg-connect-3}} \hfill
    \subfigure[]{\includegraphics[width=0.49\linewidth]{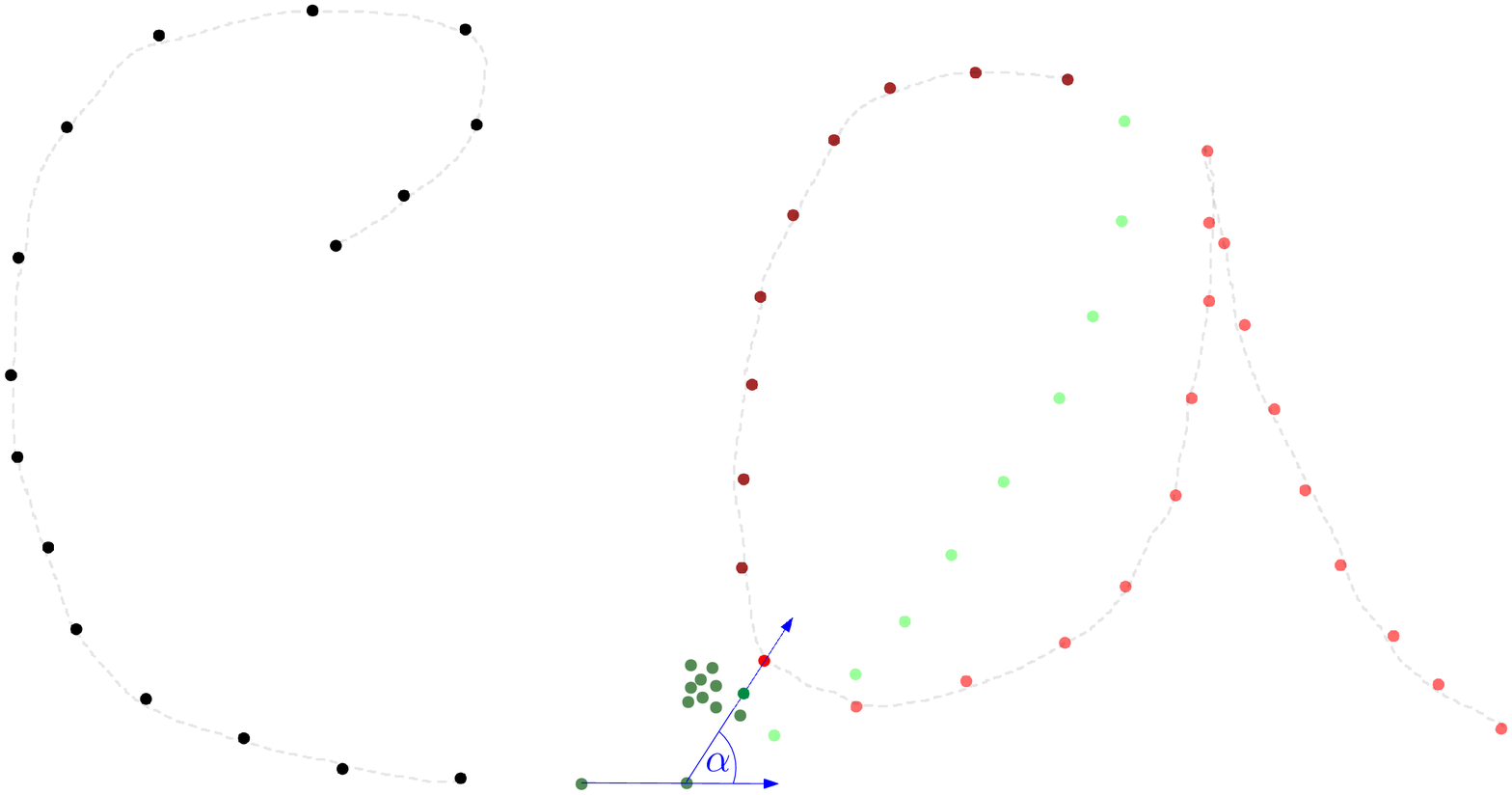} \label{fig:alg-connect-4}}
    \subfigure[]{\includegraphics[width=0.49\linewidth]{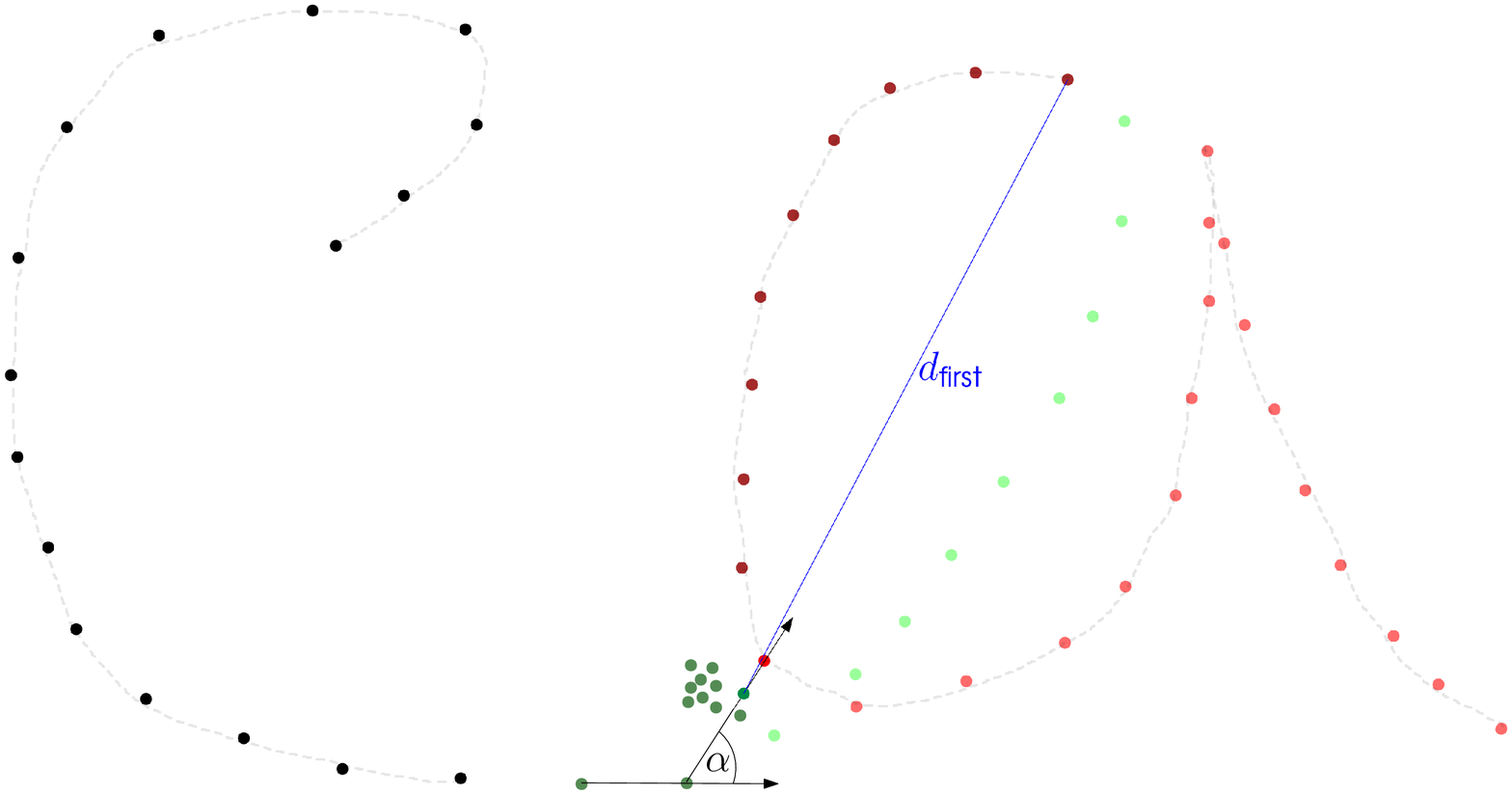} \label{fig:alg-connect-5}} \hfill
    \subfigure[]{\includegraphics[width=0.48\linewidth]{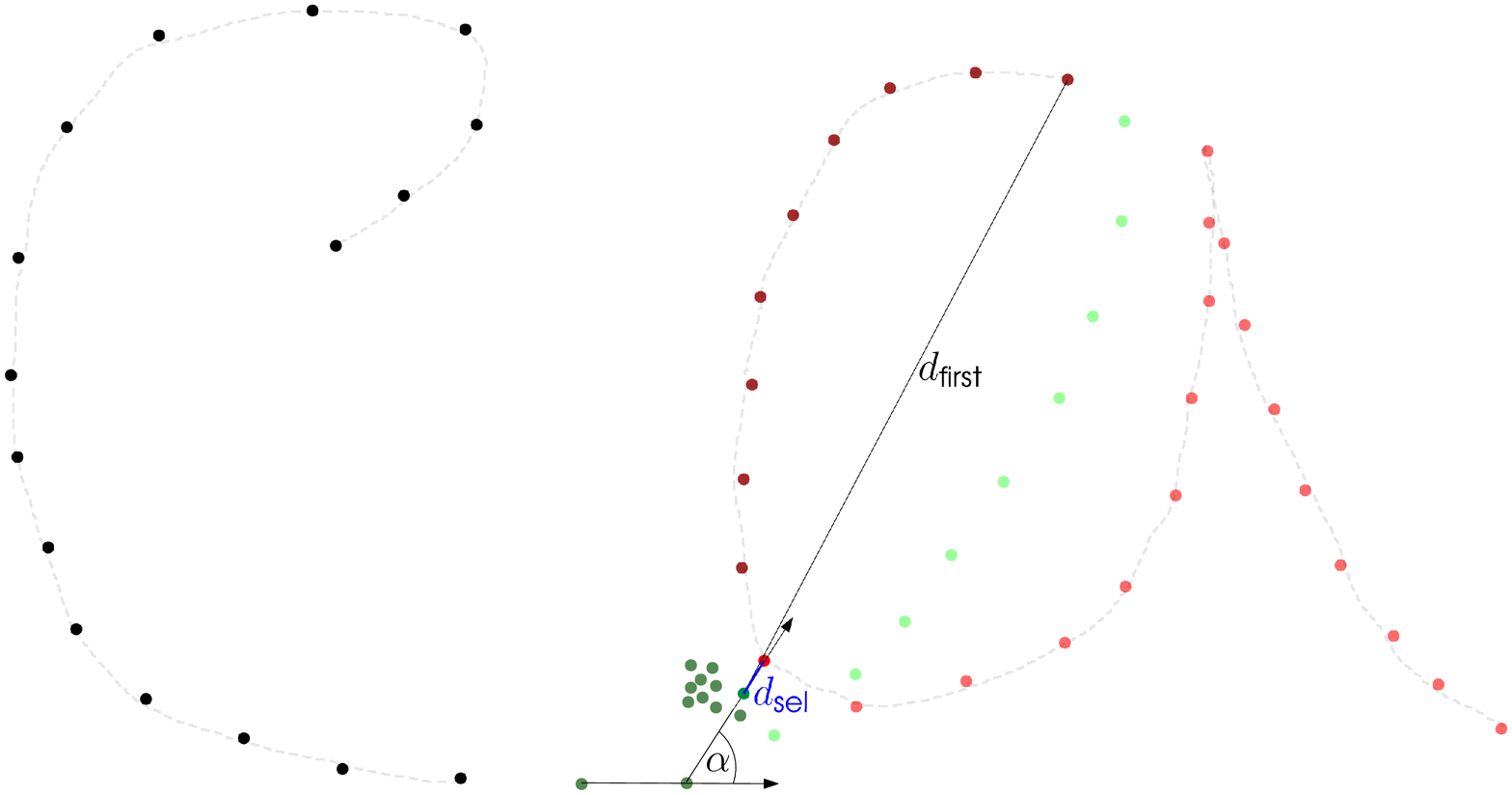} \label{fig:alg-connect-6}}
    \caption{Illustration of \Cref{alg:generation}. \subref{fig:alg-connect-1}: we present the original trajectories (black points) for two consecutive letters of a word, in this case, ``c'' and ``a''; \subref{fig:alg-connect-2}: Our algorithm will derive an intermediate trajectory to join both trajectories (dark and light green points).
    Here, \subref{fig:alg-connect-2}, we finish the second iteration since we have two computed points (dark green points) of the intermediate trajectory; \subref{fig:alg-connect-3} At each iteration, we produce a candidate point for each point of the following trajectory, in this case, ``a''.
    \subref{fig:alg-connect-4} We show the cost term to contemplate the angle between the vector composed from the last two points of the intermediate trajectory and the vector that includes the current candidate point and the point of the subsequent trajectory, ``a'', that it is trying to connect.
    \subref{fig:alg-connect-5} Cost term to consider the distance between the current candidate point and the first point of the following trajectory, ``a''. \subref{fig:alg-connect-6} Cost term to ponder the distance between the current candidate point and the point of the subsequent trajectory that it is trying to connect. Finally, the index of the same point of the following trajectory, ``a'', also weights the cost.}
    \label{fig:alg-connect}
\end{figure*}

Our newly proposed algorithm takes two consecutive trajectories (adjacent letters in a word) and infers an intermediate trajectory to connect the two. The algorithm is iterative, and at each iteration, we estimate a new point of the intermediate trajectory. At each iteration, we produce a candidate point for each point of the following letter trajectory. We compute a cost for each candidate point and select the one with the lowest cost to be the final estimation point for this iteration. The cost for each candidate point considers the angle between the vector composed from the last two points of the intermediate trajectory and the vector that includes the current candidate point and a point of the subsequent letter trajectory that it is trying to connect. This cost also has in consideration the distance between the current candidate point and the first point of the following trajectory. Another term of this cost ponders the distance between the current candidate point and the point of the subsequent trajectory that it is trying to connect. Finally, the index of the same point of the following trajectory also weights the cost. We provide the pseudo-code of our algorithm in \Cref{alg:generation} and an example of execution in \cref{fig:alg-connect}.

\begin{algorithm*}[th]
\caption{Algorithm to create an intermediate (connection) trajectory between two consecutive trajectories.}\label{alg:generation}
\begin{algorithmic}[1]
 \renewcommand{\algorithmicrequire}{\textbf{Input:}}
 \renewcommand{\algorithmicensure}{\textbf{Output:}}
 \REQUIRE {$\mathbf{x}_{T}^i$, $\mathbf{x}_{T}^{i+1}$ - Motion trajectories; \newline $\delta$ - small trajectory increment;\newline $\alpha_{\text{max}}$ - Maximum angular difference allowed between vectors;\newline $\theta_o$ - Constant for cost penalizing index of point in x;\newline $\theta_a$ - Constant for cost penalizing the angle between the last connection trajectory vector and the new candidate vector to be included;\newline $\theta_f$ - Constant for cost penalizing the distance between new candidate point and the initial point of $\mathbf{x}_{T}^{i+1}$;\newline $\theta_p$ - Constant for cost penalizing the distance between new candidate point and the point in $\mathbf{x}_{T}^{i+1}$ to connect (target).}
 \ENSURE  Trajectory, $\mathbf{x}_{\text{T}}^{i'}$, connecting $\mathbf{x}_{\text{T}}^i$ to $\mathbf{x}_{\text{T}}^{i+1}$
  \STATE $\hat{x}\gets \langle1,0\rangle$
  \STATE $N_{i} \gets length(\mathbf{x}_{\text{T}}^i)$
  \STATE $N_{i+1} \gets length(\mathbf{x}_{\text{T}}^{i+1})$
  \STATE $\mathbf{x}_\text{T}^{i'}\gets\left[\right]$
  \STATE $c \gets \mathbf{x}_{\text{T}}^i[N-1] - \mathbf{x}_{\text{T}}^i[N-2]$
  \STATE $\text{costs} \gets [0,\ldots,0],\;\text{costs} \in \mathbb{R}^{N_{i+1}}$
  \STATE $\text{angles} \gets [0,\ldots,0],\;\text{angle}s \in \mathbb{R}^{N_{i+1}}$
  \STATE $\text{cand} \gets [0,\ldots,0],\;\text{cand} \in \mathbb{R}^{N_{i+1}}$
  \REPEAT
    \FOR {\(j\gets0\) to \(N_{i+1}-1\)}
      \STATE $\alpha \gets \min\left(\alpha_{\text{max}}, \angle(\mathbf{x}_{\text{T}}^{i+1}[j], \hat{x}) - \angle(c, \hat{x}))\right)$
      \STATE $\text{angles}[j]\gets\alpha$
      \STATE $\text{cand}[j] \gets c + \delta \cdot \langle\cos\left(\alpha\right),\sqrt{ 1-cos^2\left(\alpha\right)}\rangle$
      \STATE $\text{costs}[j]\gets \theta_o \left(\frac{j}{N_{i+1}}\right) + \theta_a \left(\frac{\text{angles}[j]-\min(\text{angles})}{\max(\text{angles}) - \min(\text{angles})}\right)$
      \STATE $\text{costs}[j]\gets \text{costs}[j] + \theta_f\left(\frac{\Vert\mathbf{x}_{\text{T}}^{i+1}[0]-\text{cand}[j]\Vert-\min_k{(\Vert\mathbf{x}_{\text{T}}^{i+1}[0]-\text{cand}[k]\Vert)}}{\max_k{(\Vert\mathbf{x}_{\text{T}}^{i+1}[0]-\text{cand}[k]\Vert)}-\min_k{(\Vert\mathbf{x}_{\text{T}}^{i+1}[0]-\text{cand}[k]\Vert)}}\right)$
      \STATE $\text{costs}[j]\gets \text{costs}[j] + \theta_t\left(\frac{\Vert\mathbf{x}_{\text{T}}^{i+1}[j]-\text{cand}[j]\Vert-\min_k{(\Vert\mathbf{x}_{\text{T}}^{i+1}[k]-\text{cand}[k]\Vert)}}{\max_k{(\Vert\mathbf{x}_{\text{T}}^{i+1}[k]-\text{cand}[k]\Vert)}-\min_k{(\Vert\mathbf{x}_{\text{T}}^{i+1}[k]-\text{cand}[k]\Vert)}}\right)$
    \ENDFOR
    \STATE $j\gets \argmin\left(\text{costs}\right)$
    \STATE $c\gets \text{cand}[j]$
    \STATE $\mathbf{x}_\text{T}^{i'}\gets [\mathbf{x}_\text{T}^{i'}, c]$
  \UNTIL{\(\|\mathbf{x}_\text{T}^{i+1}[j] - c\|>\delta\)}
  \RETURN \(\mathbf{x}_\text{T}^{i'}\)
 
\end{algorithmic}
\end{algorithm*}

\end{document}